\documentclass[preprint,fleqn]{elsarticle}
\journal{}

\usepackage{amssymb,amsbsy,amsthm,amsmath,amsfonts,amscd}
\usepackage[margin=1in]{geometry}
\usepackage{bm} 
\usepackage{graphicx}
\usepackage{float}
\usepackage{multirow}
\usepackage{lineno}
\usepackage{blindtext}
\usepackage{hyperref}
\hypersetup{colorlinks, citecolor=black, filecolor=black, linkcolor=black, urlcolor=black}
\usepackage{booktabs}
\usepackage[ruled,linesnumbered]{algorithm2e}
\usepackage{appendix}
\usepackage{algpseudocode}
\usepackage{mathtools,nccmath}
\usepackage{adjustbox}
\usepackage{cleveref}
\usepackage{multicol}
\usepackage{tikz}
\usetikzlibrary{shapes, arrows, positioning}
\usepackage{soul}

\begin{document}

\begin{frontmatter}
    \title{Effective Dimensionality as an Operator Invariant for Physics-Preserving Constraint Adaptation in Physics-Informed Neural Networks}

    \author[1]{Cornelius Otchere}
    \author[1]{\texorpdfstring{Michael Shields\corref{cor}}{Michael Shields}}

    \cortext[cor]{Corresponding author. \emph{E-mail address:} \texttt{michael.shields@jhu.edu} (Michael Shields)}

    \affiliation[1]{organization={Department of Civil and Systems Engineering, Johns Hopkins University},
        addressline={Baltimore, Maryland},
        country={USA}
    }

    \begin{abstract}
        Physics-Informed Neural Networks inherently suffer from task interference because they rely on a shared parameter space to satisfy both governing differential equations and boundary conditions. We analyze this structural conflict using the Fisher Information Matrix to quantify the effective degrees of freedom ($d_{\text{eff}}$) in a physics-constrained model. Unlike the classical $d_{\text{eff}}$ which measures how many parameter directions are informed by data against a statistical prior, our $d_{\text{eff}}$ measures the dimension of the parameter directions unconstrained by the differential operator. For operators with finite-dimensional kernel, we show that $d_{\text{eff}}$ converges to the kernel dimension exactly, independent of network width, depth, or activation function, recasting it from a fit diagnostic into a structural invariant of the underlying continuous operator. For operators with infinite-dimensional kernel, $d_{\text{eff}}$ instead measures the network's finite-dimensional representational bandwidth for that kernel rather than recovering an integer invariant. Importantly, $d_{\text{eff}}$ also serves as an a priori structural diagnostic. Driving $d_\text{eff}$ of a well-posed problem to zero certifies that the physics and boundary constraints have absorbed the network's free directions. Building on this characterization, we introduce subspace projection strategies for boundary adaptation. Rather than retraining from scratch, we project parameter updates into the null space of the pre-trained physics operator so that, in exact arithmetic, new boundary conditions are satisfied without disturbing the learned physics; finite-precision drift is controlled via a predictor-corrector retraction. Gradient-based fine-tuning can reach equivalent or better boundary accuracy at full convergence, but requires significantly more wall-clock time to reach comparable boundary accuracy, together with per-problem loss weighting and optimizer scheduling. Subspace projection delivers near-equivalent quality in seconds to minutes with minimal tuning overhead. We validate the method on linear and nonlinear operators, demonstrating accurate adaptation to initial and boundary shifts and previously unencountered constraint types.
    \end{abstract}

    \begin{keyword}
        Physics-Informed Learning \sep Effective Degrees of Freedom \sep Subspace Projection
    \end{keyword}
\end{frontmatter}

\section{Introduction}

Embedding governing equations into the loss function in neural network training has had a significant impact on forward modeling~\cite{raissi2019physics,karniadakis2021physics}. Enforcing both the ordinary differential equation (ODE) or partial differential equation (PDE) and boundary conditions (BCs) simultaneously has however been documented to be computationally fragile. During training, competing objectives impose conflicting demands on the shared parameter space. Wang, Teng and Perdikaris~\cite{wang2021understanding} showed how this competition leads to gradient flow pathologies. Wang, Yu and Perdikaris~\cite{wang2022and} explained this phenomenon through the Neural Tangent kernel (NTK)~\cite{jacot2018neural} perspective and proposed strategies to mitigate them. Krishnapriyan et al.~\cite{krishnapriyan2021characterizing} also showed that the competing objectives force the model to arrive at a less optimal solution. More recently, Rathore et al.~\cite{rathore2024challenges} analyzed the PINN loss landscape and identified ill-conditioning of the loss Hessian as a fundamental source of training difficulty. Self-adaptive methods~\cite{wang2021understanding,mcclenny2023self} address this imbalance by learning task-specific loss weights during training, while gradient-surgery approaches~\cite{yu2020gradient,liu2025config} project conflicting gradients onto non-opposing directions during joint optimization. These methods reduce online interference but do not structurally decouple the competing objectives.

If a boundary condition changes, the model might have to be retrained from scratch, presenting a major bottleneck for real-world applications where boundary conditions can change frequently or be uncertain. A common solution to avoid retraining includes using hard-constraint ansatz methods. Techniques like the Theory of Functional Connections or Approximate Distance Functions eliminate BCs by composing~\cite{leake2020deep} or multiplying~\cite{lagaris1998artificial,sukumar2022exact} the network output with a function that automatically satisfies the BCs. These methods can, however, become difficult for complex boundaries and under mixed BCs. To avoid retraining, researchers have also applied transfer learning. Desai et al.~\cite{desai2022one} made boundary and initial condition adaptations on only the last layer of the network. Related transfer learning approaches adopt the same layer-partitioning heuristic~\cite{xu2023transfer,goswami2020transfer} or warm-start from pre-trained models for new BCs and parameters~\cite{wang2025transfer}. These strategies risk degrading the physics representation, as gradient updates that satisfy new boundary conditions can temporarily shift parameters away from physics-resolving configurations even when the PDE loss is retained. Operator learning~\cite{lu2021learning,li2020fourier,wang2021learning} bypasses retraining entirely by directly approximating the solution map, but requires large offline datasets spanning the space of BCs and ICs and cannot leverage a pre-trained PINN's physical representation.

The machine learning community handles the risk of catastrophic forgetting with continual learning. Algorithms with Elastic Weight Consolidation anchor parameters based on historical importance~\cite{kirkpatrick2017overcoming}. Orthogonal Weight Modification projects new updates into the null space of previous tasks to prevent cross interference~\cite{zeng2019continual,farajtabar2020orthogonal}. The scientific machine learning community has recently started to adopt these concepts. Constraint enforcement methods in Extreme Learning Machines (ELMs) achieve exact satisfaction of boundary conditions by restricting the physics solution to the admissible affine subspace~\cite{schiassi2020extreme,de2025least}, which is equivalent to projecting onto the null space of the boundary operator. Because ELMs perform projections on a static random feature space, the question of how to identify and project onto an analogous null space in a trained deep network with learned features remains open. In addition, when solving differential equations, the general solution with free parameters are found first, then the BCs are used to fit the free parameters. In practice, the physical laws governing a system are fixed while boundary conditions vary across operating scenarios, geometries, or design iterations. This motivates the need for a framework that can identify and preserve the parameter-space directions encoding the physics and adapt to new BCs without risking a catastrophic forgetting of the learned physics.

To analyze the geometry of physics-informed optimization, we leverage the spectral properties of the Fisher Information Matrix (FIM), which acts as a Riemannian metric on the parameter space~\cite{amari1998natural,karakida2019pathological}. Ansuini et al.~\cite{ansuini2019intrinsic} showed that purely data-driven models compress feature representation into manifolds of varying dimensions at deeper layers. Physics constraints reshape the parameter space differently as the network's total surviving capacity is governed by the operator, though its distribution across layers depends on architectural choices. This raises a fundamental question. How should we characterize the directions in parameter space that physics constraints leave unconstrained, and how do those directions relate to classical measures of effective degrees of freedom?

Theoretical analyses of PINN convergence~\cite{shin2020convergence,mishra2022estimates, de2022generic, doumeche2025convergence} have characterized how trained networks approximate true solutions. Our work is complementary, using the FIM spectrum to diagnose whether the network has structurally resolved the operator. Our analysis uses the FIM to define a measure of effective degrees of freedom $d_{\text{eff}}$, building on classical results in statistical learning theory~\cite{mackay1992practical, sjoberg1995overtraining}. These classical formulations define $d_{\text{eff}}$ as the dimension of parameter space that observational data identifies against a soft statistical prior. The resulting quantity captures a property of the data–prior interaction, with no canonical reference value. The physics-informed setting inverts this construction. The constraint is not a statistical belief about parameters but a differential operator with an analytical kernel of known dimension. The corresponding $d_{\text{eff}}$ measures what the operator leaves unconstrained, and its predicted value is the kernel dimension of the operator itself. The trace formula is structurally analogous, but the object being measured is dual. This duality reframes $d_{\text{eff}}$ from a fit diagnostic into a structural invariant. When computed on a trained neural network, it recovers a classical property of the underlying continuous operator that we exploit to identify the directions available for boundary adaptation.

This paper makes four contributions. First, we derive an FIM-based measure of $d_{\text{eff}}$ for physics-constrained networks and show empirically that it converges to the analytical kernel dimension of the differential operator (exactly, for finite-kernel operators), independent of network width, depth, or activation function. Second, we leverage this quantity as a structural diagnostic that detects when collocation density is insufficient to resolve the operator, exposing under-constrained training that low residual loss cannot detect. Third, building on this characterization, we develop a subspace projection method for initial and boundary adaptation that updates network parameters strictly within the operator's null space, preserving the learned physics by construction. Fourth, we introduce geometric retraction and regime anchoring techniques to control the residual drift that finite-precision arithmetic induces on the curved parameter manifold.

The rest of the paper unfolds as follows. Section~\ref{sec:math_formulation} constructs the theoretical scaffolding for $d_{\text{eff}}$. Section~\ref{sec:invariance} validates $d_{\text{eff}}$ as a physical invariant rather than a structural artifact, demonstrating its resilience against architectural variation and spectral aliasing. Section~\ref{sec:subspace_strategy} introduces the subspace projection machinery and tests it on a 1D boundary value problem. Section~\ref{sec:refinements} proposes predictor-corrector retractions and regime anchoring to control residual drift on the curved parameter manifold. Section~\ref{sec:multi_dimensional} scales to a 2D Poisson equation, contrasts subspace projection against gradient-based fine-tuning, and extends to the nonlinear Burgers equation to test adaptation under shock dynamics. Section~\ref{sec:conclusion} provides concluding remarks.

\section{Mathematical Formulation of Effective Dimensionality}
\label{sec:math_formulation}
We derive $d_{\text{eff}}$ in a statistical learning setting with observational noise. The deterministic, physics-informed regime then arises as a special case in the noiseless limit, and the statistical scaffolding clarifies the capacity interpretation of the metric.

\subsection{Asymptotic Analysis of Model Quality}
\label{asymptotic_analysis_of_model_quality}
\subsubsection{Purely Data-Driven Model}
Consider a generating model $y(x) = g(x, \theta) + e$, where $g(x, \theta)$ is a neural network and the noise term $e$ is independent and identically distributed (i.i.d.) with $\mathbb{E}[e] = 0$ and $\mathbb{E}[e^2] = \kappa_0$. The empirical risk function over a dataset $\mathcal{D} = \{x_i, y_i\}_{i=1}^N$ is defined as:
\begin{equation}
    L_N(\theta) = \frac{1}{2N} \sum_{i=1}^N (y_i - g(x_i, \theta))^2
\end{equation}
The neural network is trained by empirical risk minimization as
\begin{equation}
    \hat{\theta}_N=\arg\min_\theta L_N(\theta)
\end{equation}
Following the assumptions in Sj{\"o}berg~\cite{sjoberg1995overtraining}, this formulation constitutes a maximum likelihood estimation. We assume the estimator $\hat{\theta}_N$ converges to a "true" parameter value $\theta_0$ as $N \to \infty$, where $\theta_0$ represents a local minimum of the expected criterion function. To find the optimal parameters $\hat{\theta}_N$, the derivative of the empirical risk is set to zero and a Taylor expansion is applied around $\theta_0$:

\begin{equation}
    \begin{aligned}
        0 = \nabla L_N(\hat{\theta}_N) & \approx \nabla L_N(\theta_0) + \nabla^2L_N(\theta_0)(\hat{\theta}_N - \theta_0) \\
        \hat{\theta}_N - \theta_0      & \approx - \nabla^2L_N(\theta_0)^{+} \nabla L_N(\theta_0)
    \end{aligned}
\end{equation}

\noindent where $A^+ = (A^\mathsf{T}A)^{-1}A^\mathsf{T}$ (full column rank) or $A^+ = A^\mathsf{T}(AA^\mathsf{T})^{-1}$ (full row rank) denotes the Moore-Penrose pseudoinverse. Let $\Psi(x) = -\frac{\partial}{\partial \theta} g(x, \theta) \big|_{\theta=\theta_0}$ and $Q = \mathbb{E}_x[\Psi(x)\Psi^\mathsf{T}(x)]$. Here, $Q$ is the FIM of the model with respect to its parameters. As $N \to \infty$, the Hessian $\nabla^2 L_N(\theta_0)$ can be approximated as $Q$. This approximation stems from the expansion of the second derivative:

\begin{equation}
    \nabla^2 L_N(\theta_0) = \frac{1}{N} \sum_{i=1}^N \left(\nabla g(x_i, \theta_0)\nabla g(x_i, \theta_0)^\mathsf{T} - e_i \nabla^2 g(x_i, \theta_0)\right) \approx Q
\end{equation}

\noindent As $N \to \infty$, the second term inside the summation vanishes in expectation with respect to the noise. Consequently, the variance of the gradient over the dataset realizations is:

\begin{equation}
    \mathbb{E}_{\mathcal{D}}[\nabla L_N(\theta_0)\nabla L_N{^\mathsf{T}}(\theta_0)] = \frac{\kappa_0}{N}Q
\end{equation}

\noindent The parameter error covariance matrix $P$ is derived as

\begin{equation}
    \begin{aligned}
        P & = \mathbb{E}_{\mathcal{D}}[(\hat{\theta}_N - \theta_0)(\hat{\theta}_N - \theta_0)^\mathsf{T}] \\
          & = Q^{+} \mathbb{E}_{\mathcal{D}}[\nabla L_N(\theta_0)\nabla L_N{^\mathsf{T}}(\theta_0)] Q^{+} \\
          & =  Q^{+} \left( \frac{\kappa_0}{N} Q \right) Q^{+}                                            \\
          & = \frac{\kappa_0}{N} Q^{+}
    \end{aligned}
\end{equation}

\noindent The model quality is evaluated through the expected variance of predictions
\begin{equation}
    \begin{aligned}
        \overline{L}_N               & = \mathbb{E}_{\mathcal{D}}[\overline{L}(\hat{\theta}_N)] \\
        \overline{L}(\hat{\theta}_N) & = \mathbb{E}_{x,e}[(y-g(x,\hat{\theta}_N))^2]
    \end{aligned}
\end{equation}

\noindent Taking the Taylor expansion about $\theta_0$ yields
\begin{equation}
    \label{eq:variance_unregularized}
    \begin{aligned}
        \overline{L}_N & \approx \mathbb{E}_{\mathcal{D}}\left[ \overline{L}(\theta_0) + (\hat{\theta}_N-\theta_0)^\mathsf{T} \nabla\overline{L}(\theta_0) + \frac{1}{2}(\hat{\theta}_N-\theta_0)^{\mathsf{T}} \nabla^2\overline{L}(\theta_0)(\hat{\theta}_N-\theta_0)\right] \\
                       & = \kappa_0 + \frac{1}{2} \text{tr}\left( \mathbb{E}_{\mathcal{D}}[\nabla^2\overline{L}(\theta_0)] P \right)                                                                                                                                          \\
                       & = \kappa_0 + \frac{1}{2} \text{tr}\left( 2Q \cdot \frac{\kappa_0}{N} Q^{+} \right)                                                                                                                                                                   \\
                       & = \kappa_0 \left( 1 + \frac{\text{tr}(QQ^{+})}{N} \right)                                                                                                                                                                                            \\
                       & = \kappa_0 \left( 1 + \frac{\text{rank}(Q)}{N} \right)                                                                                                                                                                                               \\
                       & = \kappa_0 \left( 1 + \frac{r}{N} \right)
    \end{aligned}
\end{equation}

\noindent where $r = \text{rank}(Q) \leq d$ and $d$ represents the total number of parameters in the model. This result aligns with the established theory \cite{ljung1987theory}. For a fixed $N$, the model's variance scales with $r$ as the model dimensionality increases. Classically, $r$ measures how many parameter directions the data identifies (the directions consumed by noise in the unregularized setting). We reframe $r$ as the number of directions available to absorb any constraint. When a structured constraint such as a differential operator is introduced, it occupies a specific subset of these directions, and the remaining capacity defines $d_{\text{eff}}$. The next subsection makes this construction precise.

\subsubsection{Physics-Informed Model}
\label{sec:physics_informed_model}

We now introduce a physics-informed loss function that enforces both a governing operator equation constraint $\mathcal{L}g(x, \theta) = f(x)$ on the domain and a BC $\mathcal{B}g(x, \theta) = u_b(x)$ on the boundary. The total loss function for the physics-informed model is defined as:

\begin{equation}
    W_N(\theta) = L_N(\theta) + \lambda_p R_p(\theta) + \lambda_b R_b(\theta)
\end{equation}

\noindent where $R_p(\theta) = \frac{1}{2N_p} \sum_{i=1}^{N_p} (f(x_i^p) - \mathcal{L}g(x_i^p, \theta))^2$ is the PDE residual over $N_p$ collocation points, $R_b(\theta) = \frac{1}{2N_b} \sum_{j=1}^{N_b} (u_b(x_j^b) - \mathcal{B}g(x_j^b, \theta))^2$ is the boundary residual over $N_b$ boundary points, and $\lambda_p, \lambda_b$ are their respective penalty weights. The optimal parameter for this loss is found as $\tilde{\theta}_N = \arg\min_\theta W_N(\theta)$.

To analyze the asymptotic behavior of the new physics-informed estimator, we assume that the true parameter $\theta_0$ exactly satisfies the governing physical laws such that $\mathcal{L}g(x, \theta_0) = f(x)$ for all $x \in \Omega$ and $\mathcal{B}g(x, \theta_0) = u_b(x)$ for all $x \in \partial \Omega$. Consequently, the empirical residuals of both the PDE and the BCs vanish at the true parameter. This assumption is an asymptotic idealization. For finite-size networks approximating non-trivial PDE solutions, $\theta_0$ should be understood as the limit of a sequence of approximating parameters as the network capacity grows. The bias-variance analysis in Section~\ref{sec:bias_variance} addresses the finite-capacity case where this assumption is relaxed.

\noindent Taking the first derivative of the total loss $W_N(\theta)$ evaluated at $\theta_0$ gives:

\begin{equation}
    \nabla W_N(\theta_0) = \nabla L_N(\theta_0) + \lambda_p \nabla R_p(\theta_0) + \lambda_b \nabla R_b(\theta_0)
\end{equation}

\noindent Since the physical constraints are perfectly satisfied at $\theta_0$, we have $\nabla R_p(\theta_0) = 0$ and $\nabla R_b(\theta_0) = 0$. Therefore, the stochasticity in $\nabla W_N(\theta_0)$ is entirely driven by the $N$ observational data points:

\begin{equation}
    \nabla W_N(\theta_0) = \nabla L_N(\theta_0)
\end{equation}

Next, we evaluate the Hessian of $\allowbreak W_N(\theta)$ at $\theta_0$. Let $\allowbreak \Phi_p(x) = -\frac{\partial}{\partial \theta} \mathcal{L}g(x, \theta) \big|_{\theta=\theta_0}$ and $\allowbreak \Phi_b(x) = -\frac{\partial}{\partial \theta} \mathcal{B}g(x, \theta) \big|_{\theta=\theta_0}$. As $N_p, N_b \to \infty$, we can approximate the Hessians of the penalty terms as $\nabla^2 R_p(\theta_0) \approx H_p$ and $\nabla^2R_b(\theta_0) \approx H_b$, where the expected FIMs are defined over their respective domain measures:

\begin{equation}
    H_p = \mathbb{E}_{x \sim \Omega}[\Phi_p(x)\Phi_p^\mathsf{T}(x)], \quad H_b = \mathbb{E}_{x \sim \partial \Omega}[\Phi_b(x)\Phi_b^\mathsf{T}(x)]
\end{equation}

It is critical to note that $H_b$ is inherently rank-deficient. Because the boundary measure is supported strictly on a lower-dimensional manifold (reducing to a sum of Dirac measures for 1D boundary value problems), the expectation $H_b$ integrates over a severely restricted subspace. For example, in a 1D problem with two boundary points, $\text{rank}(H_b) \le 2$ regardless of the parameter dimension $d$. The expected Hessian of the combined loss is therefore:

\begin{equation}
    \nabla^2W_N(\theta_0) \approx Q + \lambda_p H_p + \lambda_b H_b = M
\end{equation}

Applying the same Taylor expansion and sandwich estimator logic used for the purely data-driven model, the parameter error covariance $\tilde{P}$ for the physics-informed model becomes:

\begin{equation}
    \begin{aligned}
        \tilde{P} & = \mathbb{E}_{\mathcal{D}}[(\tilde{\theta}_N - \theta_0)(\tilde{\theta}_N - \theta_0)^\mathsf{T}] \\
                  & = M^{+} \mathbb{E}_{\mathcal{D}}[\nabla W_N(\theta_0)\nabla W_N^\mathsf{T}(\theta_0)] M^{+}       \\
                  & = M^{+} \left( \frac{\kappa_0}{N} Q \right) M^{+}                                                 \\
                  & = \frac{\kappa_0}{N} M^{+} Q M^{+}
    \end{aligned}
\end{equation}

Finally, we evaluate the expected variance of predictions, $\overline{L}_N = \mathbb{E}_{\mathcal{D}}[\overline{L}(\tilde{\theta}_N)]$, for the new estimator. Substituting $\tilde{P}$ into our expansion of the prediction error yields:

\begin{equation}
    \label{eq:variance_regularized}
    \begin{aligned}
        \overline{L}_N & \approx \kappa_0 + \frac{1}{2} \text{tr}\left( \mathbb{E}_{\mathcal{D}}[\nabla^2\overline{L}(\theta_0)] \tilde{P} \right) \\
                       & = \kappa_0 + \frac{1}{2} \text{tr}\left( 2Q \cdot \frac{\kappa_0}{N} M^{+} Q M^{+} \right)                                \\
                       & = \kappa_0 \left( 1 + \frac{\text{tr}(Q M^{+} Q M^{+})}{N} \right)                                                        \\
                       & = \kappa_0 \left( 1 + \frac{d_{\text{eff}}(\lambda_p, \lambda_b)}{N} \right)
    \end{aligned}
\end{equation}

The \textit{effective degrees of freedom} of the physics-constrained model is defined as:
\begin{equation}
    \label{eq:deff_definition}
    d_{\text{eff}}(\lambda_p, \lambda_b) = \text{tr}\!\left( Q M^{+} Q M^{+} \right), \qquad M = Q + \lambda_p H_p + \lambda_b H_b
\end{equation}
Since $Q$, $H_p$, and $H_b$ are positive semi-definite, $M \succeq Q \succeq 0$ for any $\lambda_p, \lambda_b > 0$. This ordering guarantees $d_{\text{eff}} \le r \le d$. To see this, let $B = Q^{1/2}M^+Q^{1/2}$. By trace cycling, $d_{\text{eff}} = \text{tr}(B^2)$. $M \succeq Q \succeq 0$ implies that $M^+ \preceq Q^+$, and sandwiching by $Q^{1/2}$ gives:
\begin{equation}
    B \;\preceq\; Q^{1/2}Q^+Q^{1/2} = P_Q \;\preceq\; I
\end{equation}
where $P_Q$ is the orthogonal projector onto the range of $Q$. Since $B$ is symmetric positive semi-definite and bounded above by a projector, all its eigenvalues lie in $[0,1]$. Therefore:
\begin{equation}
    \label{eq:deff_bound}
    d_{\text{eff}} = \text{tr}(B^2) = \sum_i \sigma_i(B)^2 \;\leq\; \sum_i \sigma_i(B) = \text{tr}(B) \;\leq\; \text{tr}(P_Q) = r
\end{equation}

Eq.~\eqref{eq:variance_regularized} shows that the physics constraints act as an implicit regularizer. By reducing $d_{\text{eff}}$ relative to the unregularized capacity $r$, they lower the prediction variance without requiring additional observational data. While the regularization effect of physics constraints has been widely recognized, this derivation provides an explicitly analytical pathway linking the spectral structure of $H_p$ and $H_b$ to a quantifiable reduction in model variance.

The trace formula in Eq.~\eqref{eq:deff_definition} is structurally analogous to the effective number of parameters introduced by MacKay~\cite{mackay1992practical} and Sj{\"o}berg~\cite{sjoberg1995overtraining}. In those formulations, the constraint matrix encodes a statistical prior (a Gaussian regularizer with covariance $\alpha I$), and the resulting $d_{\text{eff}}$ measures the dimension of parameter directions in which the observational likelihood dominates the prior --- that is, the directions the data identifies. The value of $d_{\text{eff}}$ in that setting has no canonical reference. It depends jointly on data informativeness and prior strength, and is interpreted as a fit diagnostic.

Our formulation inverts this construction. The constraint matrix $H_p$ encodes a differential operator with an analytically defined kernel --- a geometric object existing independently of any data or network. The trace then measures the directions the operator leaves free, and its predicted value is set a priori by the kernel dimension of the operator. Two consequences follow. First, $d_{\text{eff}}^{\text{pde}}$, which is the effective degrees of freedom evaluated with respect to the operator equation, becomes a property of the operator rather than of the fit --- a fact we verify empirically in Section~\ref{sec:invariance}, where the quantity converges to the kernel dimension independent of network width, depth, or activation function. Second, deviations between the empirical $d_{\text{eff}}^{\text{pde}}$ and the analytical kernel dimension expose structural failures: either the network is under-resolving the operator because the collocation grid is too sparse, or the problem is ill-posed in a way that the classical statistical setting cannot detect.

The classical and physics-informed $d_{\text{eff}}$ are therefore dual quantities measured by structurally identical formulas. The duality is not cosmetic: it shifts the role of the metric from statistical diagnostic to operator invariant. While the present section provides the analytical scaffolding, Section~\ref{sec:invariance} establishes this empirically.

\subsection{Bias-Variance Tradeoff under Misspecified Physics/Model}
\label{sec:bias_variance}

The bias-variance decomposition developed in this section sits within the classical statistical interpretation of $d_{\text{eff}}$. When the model is misspecified and observational noise is present, the data-vs-prior framing of MacKay and Sj{\"o}berg remains the natural lens: the variance term tracks how observational noise propagates through the operator-constrained Hessian, and the bias term tracks the irreducible projection error from misspecification. The dual interpretation of $d_{\text{eff}}$ as an operator invariant, which we further develop in Section~\ref{geometric_interpretation} and validate in Section~\ref{sec:invariance}, applies in the deterministic forward-problem regime where observational noise vanishes and the operator itself becomes the primary object of study.

Within this classical regime, we now relax the perfect physics assumption made in Section~\ref{sec:physics_informed_model} where the physics model is either misspecified or the network is not sufficiently expressive. First, we consider the case where the true process is governed by an unknown operator, leaving a systematic residual at the optimal parameters $\theta_0$. We combine the boundary and PDE constraints into a single physics operator $\mathcal{P}g(x, \theta) = h(x)$, where $\mathcal{P}$ applies the appropriate PDE or BC based on the spatial location of $x$, and $h(x)$ is the corresponding forcing term. To unify the physical constraints, we further define a concatenated physics Hessian as $\lambda H_{\text{phys}} = \lambda_p H_p + \lambda_b H_b$. If the physics model is misspecified, then at the optimal parameters $\theta_0$, the residual of the physics constraint does not vanish.

\begin{equation}
    r(x) = h(x)-\mathcal{P}g(x, \theta_0) \neq 0
\end{equation}

\noindent Since the residual no longer vanishes, the gradient of the physics penalty $\nabla R(\theta_0)$ is non-zero. The expected gradient of the physics loss converges to:

\begin{equation}
    \lim_{N \to \infty} \nabla R(\theta_0) = \mathbb{E}_x[r(x)\Phi(x)] \equiv b
\end{equation}

\noindent where $b$ represents the structural bias vector introduced by projecting the misspecified PDE onto the parameter sensitivities $\Phi(x)$. The gradient of the total empirical risk evaluated at $\theta_0$ is therefore:

\begin{equation}
    \nabla W_N(\theta_0) \approx \nabla L_N(\theta_0) + \lambda b
\end{equation}

\noindent Applying the Taylor series expansion around $\theta_0$ and solving for the parameter error yields:

\begin{equation}
    \tilde{\theta}_N - \theta_0 \approx - [Q + \lambda H_{\text{phys}}]^{+} \left( \nabla L_N(\theta_0) + \lambda b \right)
\end{equation}

We compute the expected parameter error (bias) by taking the expectation over the dataset realizations. Since the measurement noise is zero-mean ($\mathbb{E}_{\mathcal{D}}[\nabla L_N(\theta_0)] = 0$), the expected parameter bias $\beta$ is deterministic and entirely driven by the physics misspecification:

\begin{equation}
    \beta = \mathbb{E}_{\mathcal{D}}[\tilde{\theta}_N - \theta_0] = -  \lambda (Q + \lambda H_{\text{phys}})^{+} b
\end{equation}

\noindent The total parameter error covariance matrix $\hat{P}$ now decomposes into a variance component and a squared bias component:

\begin{equation}
    \begin{aligned}
        \hat{P} & = \mathbb{E}_{\mathcal{D}}[(\tilde{\theta}_N - \theta_0)(\tilde{\theta}_N - \theta_0)^\mathsf{T}]                                                                                             \\
                & = \underbrace{\frac{\kappa_0}{N} (Q + \lambda H_{\text{phys}})^{+} Q (Q +  \lambda H_{\text{phys}})^{+}}_{\text{Variance } (\tilde{P})} + \underbrace{\beta \beta^\mathsf{T}}_{\text{Bias}^2}
    \end{aligned}
\end{equation}

\noindent Substituting this augmented covariance matrix back into the prediction error $\overline{L}_N$ yields the fundamental bias-variance decomposition for the physics-informed model.

\begin{equation}
    \label{eq:bias-variance_misspecified}
    \begin{aligned}
        \overline{L}_N & \approx \kappa_0 + \frac{1}{2} \text{tr}\left( 2Q \hat{P} \right)                                                                                               \\
                       & = \kappa_0 + \text{tr}(Q \tilde{P}) + \text{tr}(Q \beta \beta^\mathsf{T})                                                                                       \\
                       & = \kappa_0 \left(1+\frac{d_{\text{eff}}(\lambda)}{N}\right)  + \beta^\mathsf{T} Q \beta                                                                         \\
                       & = \kappa_0 \left( 1 + \frac{d_{\text{eff}}(\lambda)}{N} \right) + \lambda^2 b^\mathsf{T} (Q + \lambda H_{\text{phys}})^{+} Q (Q +\lambda H_{\text{phys}})^{+} b
    \end{aligned}
\end{equation}

Eq. \eqref{eq:bias-variance_misspecified} explicitly quantifies the bias-variance tradeoff in physics-informed machine learning. In the case where the model lacks sufficient expressivity to satisfy the physics constraint, the bias term $b$ can be interpreted as the projection of the model's representational error. In this scenario, $\theta_0$ exists in a higher dimensional space and the model is effectively trying to fit a best approximation of the true physics within its limited capacity. The bias quantifies how far this approximation is from optimal.

The bias-variance tradeoff must be carefully managed through the choice of $\lambda$. Increasing the penalty weight $\lambda$ monotonically decreases the prediction variance by reducing $d_{\text{eff}}(\lambda)$. The bias term, however, exhibits two regimes in $\lambda$. For small $\lambda$, $(Q + \lambda H_{\text{phys}})^{+} \approx Q^{+}$, and the bias grows quadratically as $\lambda^2 b^\mathsf{T} Q^{+} b = \mathcal{O}(\lambda^2)$. For large $\lambda$, $(Q + \lambda H_{\text{phys}})^{+} \approx \frac{1}{\lambda} H_{\text{phys}}^{+}$, and the $\lambda^2$ prefactor cancels against the $1/\lambda^2$ from the inverses. The bias saturates to the finite value $b^\mathsf{T} H_{\text{phys}}^{+} Q H_{\text{phys}}^{+} b$, representing the irreducible projection of the misspecification.

Optimal model performance is therefore achieved by selecting a $\lambda^*$ that balances the variance reduction of this implicit regularization against this structural bias:
\begin{equation}
    \label{eq:optimal_lamdba}
    \lambda^* = \arg\min_\lambda \left[ \kappa_0 \left( 1 + \frac{d_{\text{eff}}(\lambda)}{N} \right) + \lambda^2 b^\mathsf{T} (Q + \lambda H_{\text{phys}})^{+} Q (Q +\lambda H_{\text{phys}})^{+} b \right]
\end{equation}
The location of $\lambda^*$ depends on the relative scales of the noise floor $\kappa_0/N$ and the saturated bias $b^\mathsf{T} H_{\text{phys}}^{+} Q H_{\text{phys}}^{+} b$. When the noise dominates and the saturated bias is small, $\lambda^* \to \infty$ is optimal and strict physics enforcement is justified. When noise is small and the saturated bias is significant, $\lambda^*$ lies at finite $\lambda$ within the quadratic regime, where the marginal variance reduction balances the marginal bias growth. In practice, $b$ is often unknown and cannot be directly computed. The bias-variance decomposition is therefore primarily a theoretical tool that characterizes when physics constraints are asymptotically optimal rather than a practical recipe for selecting $\lambda$. When held-out observational data is available, cross validation can identify $\lambda^*$ empirically.

\subsection{Information-Geometric Interpretation of \texorpdfstring{$d_{\text{eff}}$}{effective dimensionality} in the Purely Forward Problem Limit}
\label{geometric_interpretation}

In the preceding sections, our analysis relied on a statistical framework where the expected prediction error is driven by measurement noise. PINNs are frequently deployed outside this framework, in a deterministic forward-problem regime where observational data is absent (or perfectly noise-free) and physical constraints are enforced strongly. In this setting, $\kappa_0 \to 0$ and $\lambda \to \infty$, taking the analysis outside the classical statistical regime in which the bias-variance decomposition was developed.

In this limit, the statistical interpretation of Eq.~\eqref{eq:variance_regularized} becomes uninformative. The prediction variance $\overline{L}_N \to 0$ as $\kappa_0 \to 0$, leaving no statistical quantity to characterize. The trace formula defining $d_{\text{eff}}$, however, remains well-defined algebraically and continues to measure a structural property of the network. To interpret this property, we pivot from a frequentist statistical perspective to an information-geometric one. Under this lens, $d_{\text{eff}}$ measures the dimension of the parameter space left unconstrained by the operator --- a quantity which, in the asymptotic limit, corresponds to the kernel dimension of the operator on function space.

The network parameterization $\theta \mapsto g(\cdot, \theta)$ associates each parameter vector with a function on $\Omega$. Even in the absence of labeled data $y$, the FIM $Q$ can be evaluated using the network's output sensitivity across the continuous spatial domain and serves as a Riemannian metric on the parameter space~\cite{amari1998natural,karakida2019pathological}. Geometrically, $Q$ captures the local curvature and representational capacity of the unconstrained network.

When physics constraints are introduced through $H_{\text{phys}}$, the geometry is restricted. In the large-$\lambda$ limit, the pseudo-inverse $(Q + \lambda H_{\text{phys}})^{+}$ vanishes on $\text{range}(H_{\text{phys}})$ and is therefore confined to $\text{null}(H_{\text{phys}})$, the parameter directions the physics leaves free. For example, when $\mathcal{L} = d^2/dx^2$ is the differential operator without boundary conditions, the 2D null space is spanned by $\{1, x\}$, and $d_{\text{eff}}^{\text{pde}}$ evaluated on a trained network recovers this analytical kernel dimension ($d_{\text{eff}}^{\text{pde}} \to 2$), independent of how the network parameterizes the solution. Section~\ref{sec:invariance} verifies this convergence across architectures. When the empirical $d_{\text{eff}}^{\text{pde}}$ matches the analytical kernel dimension of $\mathcal{L}$, the dimension of the FIM's effective null space matches the dimension of the operator's kernel. Whether the network's specific functional representation captures the correct solutions in that kernel is a separate question, addressed by classical PINN convergence theory.

\subsection{Dual Space Formulation of \texorpdfstring{$d_{\text{eff}}$}{deff}}

As formulated in Eq.~\eqref{eq:deff_definition} above, the computation of $d_{\text{eff}}$ scales as $\mathcal{O}(d^3)$ in the parameter space, making it infeasible for modern neural network architectures. Recall that $\Psi(x) = -\frac{\partial}{\partial\theta}g(x,\theta)|_{\theta=\theta_{0}}$ represents the gradient of the network output. Let $\boldsymbol{\Psi} \in \mathbb{R}^{d \times N}$ be the horizontally stacked $\Psi$ over $N$ spatial points. Under this notation, the unconstrained empirical FIM is $Q =\frac{1}{N} \boldsymbol{\Psi} \boldsymbol{\Psi}^{\mathsf{T}}$. Also, let $\boldsymbol{\Phi} \in \mathbb{R}^{d \times M}$ be the stacked Jacobian matrix of physics residuals evaluated over their respective domain points where $M=N_p+N_b$. The penalized physics Hessian is thus given by $\lambda H_{\text{phys}} = \frac{\lambda}{M} \boldsymbol{\Phi} \boldsymbol{\Phi}^T$. Substituting the definitions for the empirical FIM $Q$ and the penalized physics Hessian $\lambda H_{\text{phys}}$ into Eq.~\eqref{eq:deff_definition} yields:

\begin{equation}
    \begin{aligned}
        d_{\text{eff}} & = \text{tr}\left( \left(\frac{1}{N} \boldsymbol{\Psi} \boldsymbol{\Psi}^{\mathsf{T}}\right) \left(\frac{1}{N} \boldsymbol{\Psi} \boldsymbol{\Psi}^{\mathsf{T}} + \frac{\lambda}{M} \boldsymbol{\Phi} \boldsymbol{\Phi}^T\right)^+ \left(\frac{1}{N} \boldsymbol{\Psi} \boldsymbol{\Psi}^{\mathsf{T}}\right) \left(\frac{1}{N} \boldsymbol{\Psi} \boldsymbol{\Psi}^{\mathsf{T}} + \frac{\lambda}{M} \boldsymbol{\Phi} \boldsymbol{\Phi}^T\right)^+ \right) \\
                       & =\text{tr}\left( \left( \frac{1}{\sqrt{N}} \boldsymbol{\Psi}^{\mathsf{T}} \left(\frac{1}{N} \boldsymbol{\Psi} \boldsymbol{\Psi}^{\mathsf{T}} + \frac{\lambda}{M} \boldsymbol{\Phi} \boldsymbol{\Phi}^T\right)^+ \frac{1}{\sqrt{N}} \boldsymbol{\Psi} \right)^2 \right)
    \end{aligned}
\end{equation}
To evaluate the inner matrix term cleanly, we consolidate the parameter space by horizontally concatenating the scaled spatial and physics Jacobians into a single joint matrix $K \in \mathbb{R}^{d \times (N+M)}$:
\begin{equation}
    K = \begin{bmatrix} \frac{1}{\sqrt{N}} \boldsymbol{\Psi} & \sqrt{\frac{\lambda}{M}} \boldsymbol{\Phi} \end{bmatrix}
\end{equation}

This allows the penalized parameter-space Hessian to be cleanly factored as $K K^{\mathsf{T}} = \frac{1}{N} \boldsymbol{\Psi} \boldsymbol{\Psi}^{\mathsf{T}} + \frac{\lambda}{M} \boldsymbol{\Phi} \boldsymbol{\Phi}^T$. Substituting this factorization into our target inner term yields:
\begin{equation}
    P_{\text{core}} = \frac{1}{\sqrt{N}} \boldsymbol{\Psi}^{\mathsf{T}} (K K^{\mathsf{T}})^+ \frac{1}{\sqrt{N}} \boldsymbol{\Psi}
\end{equation}
We can reveal the global geometric structure of $P_{\text{core}}$ by embedding it into a larger block-matrix operator. If we pre-multiply $(K K^{\mathsf{T}})^+$ by $K^{\mathsf{T}}$ and post-multiply by $K$, we obtain:
\begin{equation}
    K^{\mathsf{T}} (K K^{\mathsf{T}})^+ K = \begin{bmatrix} \frac{1}{\sqrt{N}} \boldsymbol{\Psi}^{\mathsf{T}} \\ \sqrt{\frac{\lambda}{M}} \boldsymbol{\Phi}^{\mathsf{T}}\end{bmatrix} (K K^{\mathsf{T}})^+ \begin{bmatrix} \frac{1}{\sqrt{N}} \boldsymbol{\Psi} & \sqrt{\frac{\lambda}{M}} \boldsymbol{\Phi} \end{bmatrix} = \begin{bmatrix}
        \frac{1}{N} \boldsymbol{\Psi}^{\mathsf{T}} (K K^{\mathsf{T}})^+ \boldsymbol{\Psi} & * \\
        *                                                                                 & *
    \end{bmatrix}
\end{equation}
where * denotes entries that are not of interest. This expansion identifies $P_{\text{core}}$ as the top-left $N \times N$ block of the global operator $K^{\mathsf{T}} (K K^{\mathsf{T}})^+ K$. This operator represents the unique orthogonal projection onto the range space of $K^{\mathsf{T}}$.

We can now cleanly transition to the dual function space. A fundamental identity of the Moore-Penrose pseudo-inverse states that the projection onto the range space of $K^{\mathsf{T}}$ can be equivalently computed using its Gram matrix:
\begin{equation}
    K^{\mathsf{T}} (K K^{\mathsf{T}})^+ K = (K^{\mathsf{T}} K) (K^{\mathsf{T}} K)^+
\end{equation}

This identity bridges the parameter space and the function space. Rather than computing the projection onto the range space of $K^{\mathsf{T}}$ via $(K K^{\mathsf{T}})^+$ in the d-dimensional parameter space, we compute it using the dual Gram matrix $K^{\mathsf{T}} K$, which is the physics-augmented empirical NTK, $G \in \mathbb{R}^{(N+M) \times (N+M)}$:
\begin{equation}
    G = K^{\mathsf{T}} K = \begin{bmatrix} \frac{1}{\sqrt{N}} \boldsymbol{\Psi}^{\mathsf{T}} \\ \sqrt{\frac{\lambda}{M}} \boldsymbol{\Phi}^{\mathsf{T}}\end{bmatrix} \begin{bmatrix} \frac{1}{\sqrt{N}} \boldsymbol{\Psi} & \sqrt{\frac{\lambda}{M}} \boldsymbol{\Phi} \end{bmatrix} = \begin{bmatrix}
        \frac{1}{N} \boldsymbol{\Psi}^{\mathsf{T}} \boldsymbol{\Psi}                     & \frac{\sqrt{\lambda}}{\sqrt{NM}} \boldsymbol{\Psi}^{\mathsf{T}} \boldsymbol{\Phi} \\
        \frac{\sqrt{\lambda}}{\sqrt{NM}} \boldsymbol{\Phi}^{\mathsf{T}}\boldsymbol{\Psi} & \frac{\lambda}{M} \boldsymbol{\Phi}^{\mathsf{T}}\boldsymbol{\Phi}
    \end{bmatrix}
\end{equation}

Because of the projection identity, the global orthogonal projection matrix is identically given by the dual projector $\Pi_G = G G^+$. Since our target term $P_{\text{core}}$ was established as the top-left block of $K^{\mathsf{T}} (K K^{\mathsf{T}})^+ K$, it must identically be the top-left $N \times N$ block of $G G^+$. Let us denote this dual block as $P_{11}$. The effective dimension is therefore $d_{\text{eff}} = \text{tr}(P_{11}^2)$ computed entirely in the (N+M)-dimensional function space. This strategy reduces the computational complexity of evaluating $d_{\text{eff}}$ from $\mathcal{O}(d^3)$ to $\mathcal{O}((N+M)^3)$, which is tractable for modern neural networks where $(N+M) \ll d$.

\section{Invariance and Resolution of \texorpdfstring{$d_{\text{eff}}$}{deff}}
\label{sec:invariance}

All experiments in this paper were performed on a single NVIDIA A100 GPU.

\subsection{Invariance for Linear Operators}

A fundamental criticism of dimensionality metrics in deep learning is their sensitivity to architectural bloat~\cite{ansuini2019intrinsic}. If a capacity metric merely tracks parameter counts, it cannot reliably diagnose the physical well-posedness of a continuous boundary value problem. Let $d_{\text{eff}}^{\text{pde}}$ represent the dimensionality of the subspace that is unconstrained by the operator equation. We hypothesize that if $d_{\text{eff}}^{\text{pde}}$ accurately measures the surviving degrees of freedom of a physics operator, it must converge to the dimension of analytical kernel of the differential equation, entirely independent of the underlying neural architecture.

We test this by conducting a parametric sweep across three governing equations: a 1st-order, 2nd-order, and 3rd-order ordinary differential equation (ODE). The specific operators used for this sweep are:
\begin{align}
    \text{1st-order:} \quad & \frac{du}{dx} = 2\pi\cos(2\pi x)  \quad x \in [-1, 1] \label{eq:ode1}         \\
    \text{2nd-order:} \quad & \frac{d^2u}{dx^2} = -4\pi^2 \sin(2\pi x)  \quad x \in [-1, 1] \label{eq:ode2} \\
    \text{3rd-order:} \quad & \frac{d^3u}{dx^3} = -8\pi^3 \cos(2\pi x) \quad x \in [-1, 1] \label{eq:ode3}
\end{align}
Analytically, integrating an $n$-th order ODE yields exactly $n$ free constants, so the solution manifold should be $n$-dimensional. We therefore expect $d_{\text{eff}}^{\text{pde}}\to n$ for each operator.

\begin{table}[htbp!]
    \centering
    \caption{Conservation of capacity ($d_{\text{eff}}^{\text{pde}}$) across architectures. Feedforward = standard MLP and Skip Connections = residual architecture with skip connections injected into hidden layers.}
    \label{tab:capacity_conservation}
    \resizebox{\textwidth}{!}{
        \begin{tabular}{cccccccccc}
            \toprule
                                  &                &                     &                        &                & \multicolumn{5}{c}{$d_{\text{eff}}^{\text{pde}}$}                                                                             \\
            \cmidrule(lr){6-10}
            \textbf{Architecture} & \textbf{Order} & \textbf{Activation} & \textbf{Hidden Layers} & \textbf{Width} & \textbf{All Layers}                               & \textbf{Layer 0} & \textbf{Layer 1} & \textbf{Layer 2} & \textbf{Layer 3} \\
            \midrule
            Feedforward           & 1              & SiLU                & 1                      & 50             & 1.098                                             & 1.098            & 1.097            & -                & -                \\
            Feedforward           & 1              & Tanh                & 1                      & 50             & 1.098                                             & 1.098            & 1.098            & -                & -                \\
            Feedforward           & 1              & SiLU                & 3                      & 100            & 1.098                                             & 1.098            & 1.098            & 1.098            & 1.098            \\
            Feedforward           & 1              & Tanh                & 3                      & 100            & 1.098                                             & 1.098            & 1.098            & 1.098            & 1.098            \\
            \midrule
            Feedforward           & 2              & SiLU                & 1                      & 50             & 2.001                                             & 2.001            & 2.001            & -                & -                \\
            Feedforward           & 2              & Tanh                & 1                      & 50             & 2.001                                             & 2.001            & 2.001            & -                & -                \\
            Feedforward           & 2              & SiLU                & 3                      & 100            & 2.001                                             & 2.001            & 2.001            & 2.001            & 2.001            \\
            Feedforward           & 2              & Tanh                & 3                      & 100            & 2.001                                             & 2.001            & 2.001            & 2.001            & 2.001            \\
            \midrule
            Feedforward           & 3              & SiLU                & 1                      & 50             & 3.000                                             & 3.000            & 3.000            & -                & -                \\
            Feedforward           & 3              & Tanh                & 1                      & 50             & 3.000                                             & 3.000            & 3.000            & -                & -                \\
            Feedforward           & 3              & SiLU                & 3                      & 100            & 3.000                                             & 3.000            & 3.000            & 3.000            & 3.000            \\
            Feedforward           & 3              & Tanh                & 3                      & 100            & 3.000                                             & 2.757            & 2.999            & 2.999            & 3.000            \\
            \midrule
            Skip Connections      & 1              & SiLU                & 1                      & 50             & 1.098                                             & 1.098            & 1.097            & -                & -                \\
            Skip Connections      & 1              & Tanh                & 1                      & 50             & 1.098                                             & 1.098            & 1.098            & -                & -                \\
            Skip Connections      & 1              & SiLU                & 3                      & 100            & 1.098                                             & 1.098            & 1.098            & 1.098            & 1.098            \\
            Skip Connections      & 1              & Tanh                & 3                      & 100            & 1.098                                             & 1.098            & 1.098            & 1.098            & 1.098            \\
            \midrule
            Skip Connections      & 2              & SiLU                & 1                      & 50             & 2.001                                             & 2.001            & 2.001            & -                & -                \\
            Skip Connections      & 2              & Tanh                & 1                      & 50             & 2.001                                             & 2.001            & 2.001            & -                & -                \\
            Skip Connections      & 2              & SiLU                & 3                      & 100            & 2.001                                             & 2.001            & 2.001            & 2.001            & 2.001            \\
            Skip Connections      & 2              & Tanh                & 3                      & 100            & 2.001                                             & 2.001            & 2.001            & 2.001            & 2.001            \\
            \midrule
            Skip Connections      & 3              & SiLU                & 1                      & 50             & 3.000                                             & 3.000            & 3.000            & -                & -                \\
            Skip Connections      & 3              & Tanh                & 1                      & 50             & 3.000                                             & 3.000            & 3.000            & -                & -                \\
            Skip Connections      & 3              & SiLU                & 3                      & 100            & 3.000                                             & 3.000            & 3.000            & 3.000            & 3.000            \\
            Skip Connections      & 3              & Tanh                & 3                      & 100            & 3.000                                             & 3.000            & 3.000            & 3.000            & 3.000            \\
            \bottomrule
        \end{tabular}
    }
\end{table}

For 1D experiments, all pseudoinverse computations use rcond $= 10^{-10}$, where singular values smaller than rcond $\times$ largest singular value are treated as zero. While the invariance results hold up to an rcond of $10^{-15}$, we chose $10^{-10}$ to avoid retaining numerically negligible singular values that carry no physical significance during the adaptation introduced in Section~\ref{sec:subspace_strategy}. The results in Table \ref{tab:capacity_conservation} demonstrate the independence of the metric from the architecture. For the 2nd-order ODE, despite parameter counts ranging from 150 to over 30,000, the effective dimension rigidly flatlines at $d_{\text{eff}} \approx 2.00$. The metric successfully identifies the geometric space not dictated by the physics while ignoring thousands of redundant parameters. The 1st-order system exhibits similar architecture-invariance pattern, locking onto $d_{\text{eff}}^{\text{pde}} \approx 1.098$. The offset above the analytical value $1$ is small and stable. It decreases marginally with collocation density ($1.0975 \text{ at }N=100, 1.0949 \text{ at }N=1000, 1.0947 \text{ at }N=10000$).

The sweep also exposes a critical distinction between local and global tangent spaces. In the deeper 3rd-order feedforward networks, individual layer capacities can bottleneck. The input layer of the 3-layer Tanh feedforward network drops to $d_{\text{eff}}^{\text{local}} \approx 2.757$, below the analytical value of 3. Yet Layers 1, 2, and 3 each recover to $\approx 2.999$, and evaluating the entire parameter space simultaneously yields a global capacity of exactly $d_{\text{eff}}^{\text{global}} = 3.000$. The global Jacobian is the horizontal concatenation of the local Jacobians. The global capacity is preserved because layers compensate for each other. When one layer's local null space cannot fully span the operator kernel, subsequent layers absorb the unmet constraint. The PDE operator requires only the union of layer-wise tangent spaces to construct the full solution manifold.

The per-layer picture exposes a structural risk: a layer whose local $d_{\text{eff}}$ falls below the operator's kernel dimension lacks sufficient capacity to independently resolve all null-space directions. The skip connection rows in Table \ref{tab:capacity_conservation} confirm the global $d_{\text{eff}} = N$ invariant while showing that skip connections~\cite{he2016deep} can eliminate the early-layer bottleneck. In the 3-layer Tanh skip connection network, $d_{\text{eff}}^{\text{local}}$ at Layer 0 recovers from $\approx 2.757$ to exactly $3.000$. Skip connections allow gradient paths to bypass constrained neurons, preventing any single layer from being disproportionately consumed by the PDE constraint. This distributes capacity evenly across depth and preserves Layer 0 as a viable BC injection site.

A natural reading is that $d_{\text{eff}}^{\text{pde}}$ simply
reports the order of the differential operator. The 2D Poisson operator
separates the two interpretations and is taken up in
Section~\ref{sec:2d_poisson}. We note here only that a second-order operator
with an infinite-dimensional kernel does not return a small finite integer,
which an order-counting metric would require.

The null spaces of the ODEs in Eqs.\eqref{eq:ode1}--\eqref{eq:ode3} are spanned by elementary functions. A sharper test is an operator where the null space is no longer elementary. The order-zero Bessel equation,
\begin{equation}
    \label{eq:bessel}
    x^2 \frac{d^2u}{dx^2} + x \frac{du}{dx} + x^2 u = 0, \qquad x \in [1, 8],
\end{equation}
has a two-dimensional kernel spanned by the Bessel functions $J_0(x)$ and $Y_0(x)$. Again, we hypothesize that $d_{\text{eff}}^{\text{pde}}$ converges to 2 regardless of architecture. Because the equation is homogeneous, training without boundary conditions can admit the trivial solution $u \equiv 0$, which yields a degenerate Jacobian and a meaningless $d_{\text{eff}}$. We therefore anchor each network to the non-trivial solution $J_0(x)$ by imposing two Dirichlet conditions, $u(1) = J_0(1) \approx 0.7652 \text{, and }u(8) = J_0(8) \approx 0.1717$.

\begin{table}[htbp!]
    \centering
    \caption{Conservation of $d_{\text{eff}}^{\text{pde}}$ for the Bessel operator~\eqref{eq:bessel}.}
    \label{tab:bessel_deff}
    \resizebox{\textwidth}{!}{
        \begin{tabular}{ccccccccc}
            \toprule
                           &                     &                        &                  & \multicolumn{5}{c}{$d_{\text{eff}}^{\text{pde}}$}                                 \\
            \cmidrule(lr){5-9}
            \textbf{Model} & \textbf{Activation} & \textbf{Hidden Layers} & \textbf{Width}
                           & \textbf{All Layers} & \textbf{Layer 0}       & \textbf{Layer 1}
                           & \textbf{Layer 2}    & \textbf{Layer 3}                                                                                                              \\
            \midrule
            \multirow{4}{*}{MLP}
                           & SiLU                & 1                      & 50               & 1.998                                             & 2.000 & 1.942 & -     & -     \\
                           & Tanh                & 1                      & 50               & 1.999                                             & 1.998 & 1.831 & -     & -     \\
                           & SiLU                & 3                      & 100              & 2.001                                             & 2.001 & 2.001 & 2.001 & 2.001 \\
                           & Tanh                & 3                      & 100              & 2.000                                             & 2.000 & 2.000 & 1.997 & 1.982 \\
            \midrule
            \multirow{4}{*}{ResNet}
                           & SiLU                & 1                      & 50               & 1.985                                             & 1.984 & 1.764 & -     & -     \\
                           & Tanh                & 1                      & 50               & 1.998                                             & 1.998 & 1.742 & -     & -     \\
                           & SiLU                & 3                      & 100              & 2.000                                             & 1.998 & 2.000 & 2.000 & 1.989 \\
                           & Tanh                & 3                      & 100              & 2.000                                             & 2.000 & 1.997 & 1.993 & 1.982 \\
            \bottomrule
        \end{tabular}
    }
\end{table}

Table~\ref{tab:bessel_deff} confirms the invariant. Across all eight architecture/activation combinations, the global $d_{\text{eff}}^{\text{pde}}$ converges to $\approx 2$, matching the analytical kernel dimension of~\eqref{eq:bessel}. The global-vs-local breakdown reveals minimal deviations from the expected value.

\subsection{Invariance for Nonlinear Operators}

The experiments so far have used linear differential operators. In this section, we extend the experiments to nonlinear operators. Nonlinear operators do not have a kernel in the same sense as linear operators but a local solution manifold of admissible perturbations. To test whether $d_{\text{eff}}^{\text{pde}}$ is invariant in the nonlinear setting, we evaluate two second-order nonlinear operators, both admitting the same exact solution $u^* = \sin(\pi x)$ on $[-1, 1]$ with Dirichlet conditions $u(-1) = u(1) = 0$:
\begin{align}
    \frac{d^2u}{dx^2} + u\,\frac{du}{dx} & = -\pi^2\sin(\pi x) + \frac{\pi}{2}\sin(2\pi x) \label{eq:forced_burgers} \\
    \frac{d^2u}{dx^2} + u^2              & = -\pi^2\sin(\pi x) + \sin^2(\pi x) \label{eq:quadratic_nl}
\end{align}

\begin{table}[htbp!]
    \centering
    \caption{$d_{\text{eff}}^{\text{pde}}$ for two second-order nonlinear operators (Eqs.~\eqref{eq:forced_burgers} and~\eqref{eq:quadratic_nl}).}
    \label{tab:nonlinear_deff}
    \resizebox{\textwidth}{!}{
        \begin{tabular}{llcccccccc}
            \toprule
                              &                       &                     &                        &                & \multicolumn{5}{c}{$d_{\text{eff}}^{\text{pde}}$}                                                                             \\
            \cmidrule(lr){6-10}
            \textbf{Equation} & \textbf{Architecture} & \textbf{Activation} & \textbf{Hidden Layers} & \textbf{Width} & \textbf{All Layers}                               & \textbf{Layer 0} & \textbf{Layer 1} & \textbf{Layer 2} & \textbf{Layer 3} \\
            \midrule
            \multirow{8}{*}{\eqref{eq:forced_burgers}}
                              & Feedforward           & SiLU                & 1                      & 50             & 2.001                                             & 2.001            & 2.001            & -                & -                \\
                              & Feedforward           & Tanh                & 1                      & 50             & 2.001                                             & 2.001            & 2.001            & -                & -                \\
                              & Feedforward           & SiLU                & 3                      & 100            & 2.001                                             & 2.001            & 2.001            & 2.001            & 2.001            \\
                              & Feedforward           & Tanh                & 3                      & 100            & 2.001                                             & 2.001            & 2.001            & 2.001            & 2.001            \\
                              & Skip Connections      & SiLU                & 1                      & 50             & 2.001                                             & 2.001            & 2.001            & -                & -                \\
                              & Skip Connections      & Tanh                & 1                      & 50             & 2.001                                             & 2.001            & 2.001            & -                & -                \\
                              & Skip Connections      & SiLU                & 3                      & 100            & 2.001                                             & 2.001            & 2.001            & 2.001            & 2.001            \\
                              & Skip Connections      & Tanh                & 3                      & 100            & 2.001                                             & 2.001            & 2.001            & 2.001            & 2.001            \\
            \midrule
            \multirow{8}{*}{\eqref{eq:quadratic_nl}}
                              & Feedforward           & SiLU                & 1                      & 50             & 2.002                                             & 2.002            & 2.002            & -                & -                \\
                              & Feedforward           & Tanh                & 1                      & 50             & 2.001                                             & 2.001            & 2.001            & -                & -                \\
                              & Feedforward           & SiLU                & 3                      & 100            & 2.001                                             & 2.001            & 2.001            & 2.001            & 2.001            \\
                              & Feedforward           & Tanh                & 3                      & 100            & 2.001                                             & 2.001            & 2.001            & 2.001            & 2.001            \\
                              & Skip Connections      & SiLU                & 1                      & 50             & 2.002                                             & 2.002            & 2.002            & -                & -                \\
                              & Skip Connections      & Tanh                & 1                      & 50             & 2.001                                             & 2.001            & 2.001            & -                & -                \\
                              & Skip Connections      & SiLU                & 3                      & 100            & 2.002                                             & 2.002            & 2.002            & 2.002            & 2.002            \\
                              & Skip Connections      & Tanh                & 3                      & 100            & 2.001                                             & 2.001            & 2.001            & 2.001            & 2.001            \\
            \bottomrule
        \end{tabular}
    }
\end{table}

Table~\ref{tab:nonlinear_deff} shows that $d_{\text{eff}}^{\text{pde}} \approx 2.00$ across all sixteen configurations for both equations. The small residual offsets are consistent with the finite-grid discretization artifact observed for the linear operators in Section~\ref{sec:invariance}.

\subsection{The Spatial Sieve and the Illusion of Interpolation}

In classical supervised learning, model validity is established via empirical test error. Physics-informed neural networks aim to establish model validity without relying on solutions we do not possess. Validity must therefore be established structurally without ground truth. Relying on isolated training points can be deceptive because a network can fit the residual to machine precision at the collocation points while producing either wild oscillations or flat regions between them. We require a structural diagnostic that detects under-resolution without appealing to ground truth. Continuous PDE residual evaluation on a dense grid remains a valuable tool, but it does not provide any structural information. $d_{\text{eff}}$ exposes how many unconstrained directions remain and, when computed layerwise, where in the network they reside.

We isolate this phenomenon using the 1D 2nd-order ODE in Eq.~\eqref{eq:ode2} subject to strict Dirichlet boundaries ($u(-1)=u(1)=1$). This BVP has a unique analytical solution. A fully resolved system should consume the network's representation capacity, leaving zero unconstrained degrees of freedom. We train an expressive network (4 hidden layers, 100 neurons per layer, SiLU activation) against progressively denser spatial grids and track $d_{\text{eff}}^{\text{total}}$ before and after optimization as well as $d_{\text{eff}}^{\text{pde}}$. The total effective dimension is computed from the stacked Jacobian $[\mathbf{J}_{\text{pde}}\, \sqrt{\alpha}\,\mathbf{J}_{\text{bc}}]$, where $\alpha = \sigma_{\max}(\mathbf{J}_{\text{pde}})^2 / \sigma_{\max}(\mathbf{J}_{\text{bc}})^2$ spectrally balances the two blocks so that neither the PDE nor the BC constraints dominate the combined pseudoinverse.

\begin{figure}[htbp!]
    \centering
    \includegraphics[width=\textwidth]{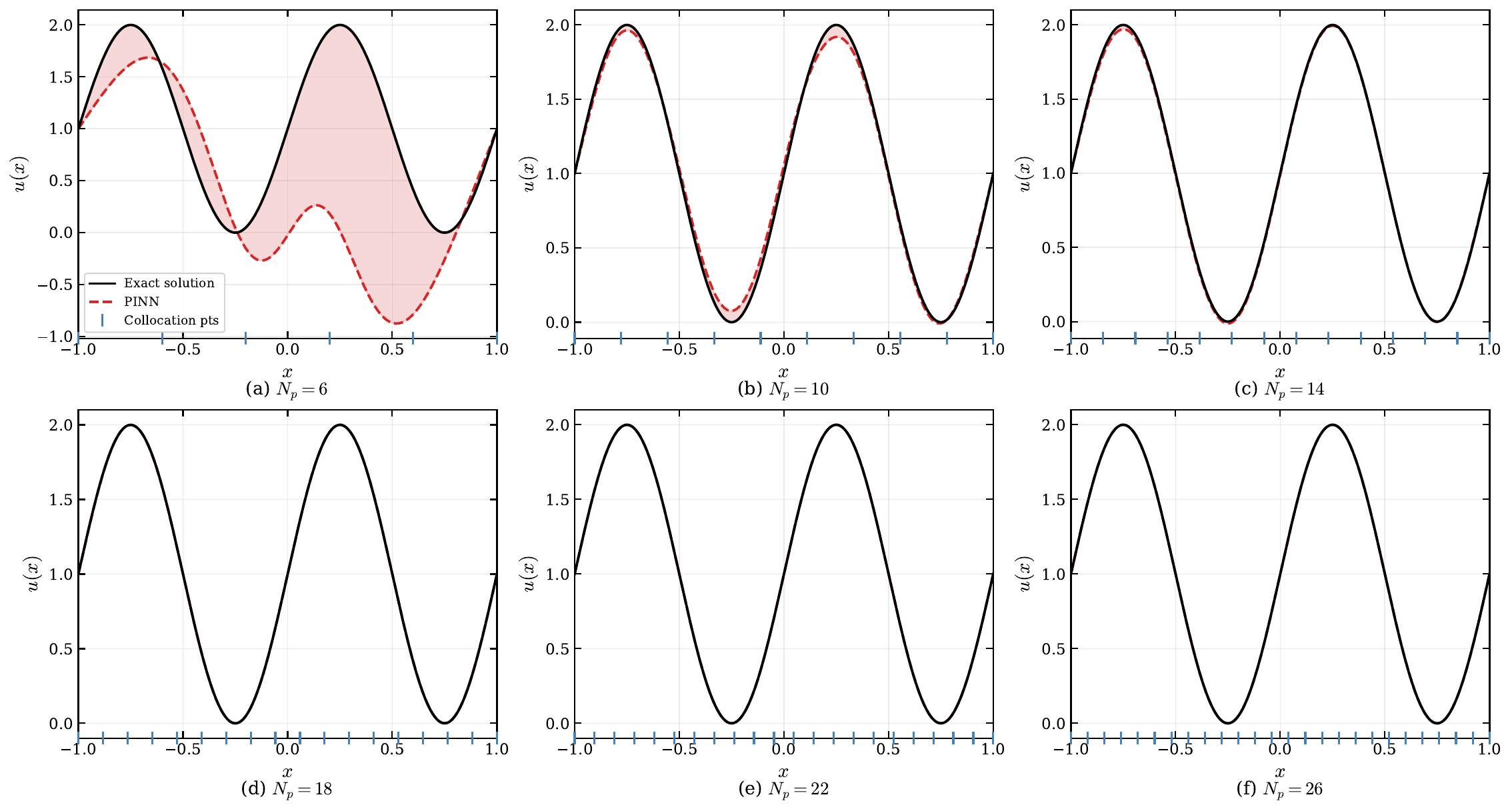}
    \caption{Spatial sieve experiment with the feedforward architecture. Each panel shows the exact solution (black) and the PINN prediction (red dashed) with shaded error. Blue tick marks at the bottom of each panel indicate the collocation points. As the grid density increases from left to right and top to bottom, the continuous solution converges toward the exact answer.}
    \label{fig:spatial_sieve_feedforward}
\end{figure}

\begin{table}[htbp!]
    \centering
    \caption{The Spatial Sieve across architectures.}
    \label{tab:spatial_sieve}
    \resizebox{\textwidth}{!}{
        \begin{tabular}{lccccccc}
            \toprule
            \textbf{Architecture}            & \textbf{\shortstack{Grid                                                                                                 \\Points}} & \textbf{\shortstack{Pre-Train                                                                                      \\$d_{\text{eff}}^{\text{total}}$}} & \textbf{\shortstack{Post-Train                                                                                      \\$d_{\text{eff}}^{\text{total}}$}} & \textbf{\shortstack{Post-Train                                                                                      \\$d_{\text{eff}}^{\text{pde}}$}} & \textbf{\shortstack{Discrete \\PDE Residual}} & \textbf{\shortstack{Continuous \\PDE Residual}} & \textbf{\shortstack{Continuous \\ RMSE}}\\
            \midrule
            \multirow{6}{*}{Feedforward}     & 6                        & 1.950 & 4.000  & 6.000  & $1.69\times 10^{-29}$ & $5.28\times 10^{2}$  & $1.05\times 10^{0}$  \\
                                             & 10                       & 1.940 & 8.010  & 10.000 & $1.11\times 10^{-28}$ & $1.10\times 10^{1}$  & $4.49\times 10^{-2}$ \\
                                             & 14                       & 1.940 & 10.000 & 12.000 & $1.05\times 10^{-28}$ & $9.16\times 10^{-1}$ & $1.32\times 10^{-2}$ \\
                                             & 18                       & 1.940 & 3.000  & 6.000  & $2.12\times 10^{-17}$ & $1.96\times 10^{-1}$ & $1.20\times 10^{-3}$ \\
                                             & 22                       & 1.930 & 0.107  & 2.000  & $9.67\times 10^{-18}$ & $3.77\times 10^{-2}$ & $3.60\times 10^{-4}$ \\
                                             & 26                       & 1.930 & 0.103  & 2.000  & $3.63\times 10^{-7}$  & $1.11\times 10^{-3}$ & $7.36\times 10^{-5}$ \\
            \midrule
            \multirow{6}{*}{Skip Connection} & 6                        & 0.765 & 4.000  & 6.000  & $1.05\times 10^{-29}$ & $1.80\times 10^{2}$  & $3.58\times 10^{-1}$ \\
                                             & 10                       & 0.724 & 8.000  & 10.000 & $1.40\times 10^{-29}$ & $1.13\times 10^{1}$  & $2.11\times 10^{-2}$ \\
                                             & 14                       & 0.708 & 7.000  & 9.000  & $4.42\times 10^{-24}$ & $6.80\times 10^{-1}$ & $4.04\times 10^{-3}$ \\
                                             & 18                       & 0.699 & 3.000  & 4.000  & $4.75\times 10^{-17}$ & $2.06\times 10^{-1}$ & $1.60\times 10^{-3}$ \\
                                             & 22                       & 0.694 & 0.047  & 2.000  & $1.34\times 10^{-16}$ & $2.74\times 10^{-2}$ & $4.00\times 10^{-4}$ \\
                                             & 26                       & 0.690 & 0.046  & 2.000  & $6.13\times 10^{-7}$  & $1.68\times 10^{-3}$ & $9.41\times 10^{-5}$ \\
            \bottomrule
        \end{tabular}
    }
\end{table}

Table \ref{tab:spatial_sieve} reveals the transition between under-resolved and resolved regimes. Before optimization, $d_{\text{eff}}^{\text{total}}\approx 1.94$ for the feedforward network and $\approx 0.71$ for the skip-connection network, highlighting the architectural difference. While the pre-training capacity is architecture-determined, both architectures exhibit similar spatial sieve behavior post-training. At 6 grid points, the optimizer drives the discrete PDE and BC residuals to machine precision. Relying on training loss alone would falsely indicate convergence. The post-training metric exposes the illusion: with only 6 collocation points, the physics Jacobian has rank at most 6, so $d_{\text{eff}}^{\text{pde}} \leq 6$ regardless of the operator. The observed value $d_{\text{eff}}^{\text{pde}} = 6.00$ saturates this ceiling, indicating that none of the available constraint capacity has been redirected by the operator structure. The remaining gap of two between $d_{\text{eff}}^{\text{pde}}$ and $d_{\text{eff}}^{\text{total}} \approx 4.00$ reflects the two boundary conditions adding additional constraints. This structural under-resolution is verified directly by the continuous PDE residual ($5.28 \times 10^{2}$) and continuous RMSE ($1.05$), both of which would be small if the network had actually solved the operator problem. Figure~\ref{fig:spatial_sieve_feedforward}(a) visually confirms the under-resolution, showing that the network has not learned the correct solution despite zero training loss. At 10 points, both architectures remain structurally unresolved: $d_{\text{eff}}^{\text{pde}} = 10.00$ saturates the rank ceiling of 10 for both, with $d_{\text{eff}}^{\text{total}} \approx 8.00$ reflecting the two-BC gap. As the spatial sieve tightens further, $d_{\text{eff}}^{\text{pde}}$ begins to fall below the rank ceiling. At 14 points, the feedforward network reads $d_{\text{eff}}^{\text{pde}} \approx 12.00$ (near the ceiling of 14), still largely saturated, while the skip-connection network drops to $\approx 9.00$, indicating an earlier onset of operator constraint. At 18 points the feedforward reads $d_{\text{eff}}^{\text{pde}} \approx 6.00$ and the skip connection $\approx 4.00$. By 22 points, both architectures fully resolve the operator: $d_{\text{eff}}^{\text{pde}} \approx 2.00$ matches the analytical kernel dimension and $d_{\text{eff}}^{\text{total}} \approx 0.1$ confirms the BCs have absorbed the remaining capacity.

Driving $d_{\text{eff}}^{\text{total}} \to 0$ is a necessary, but not sufficient, condition for physical accuracy. A collapsed capacity certifies that no unconstrained directions remain at the converged parameters. When combined with the well-posedness of the underlying PDE/BC system and convergence to the correct attractor basin, this implies local uniqueness of the network's solution. The network itself may still possess parameter-space gauge freedoms (directions that leave the function output unchanged), but these are invisible to $d_{\text{eff}}^{\text{total}}$ and irrelevant to physical correctness. The diagnostic does not, however, certify that the optimizer has found the correct attractor. If the boundary conditions are misspecified, or if the optimizer becomes trapped in a pathological local minimum, the manifold can structurally lock into an incorrect geometric state with $d_{\text{eff}}^{\text{total}} \to 0$ still satisfied.

\section{Boundary Adaptation via Subspace Projection}
\label{sec:subspace_strategy}

Section~\ref{sec:invariance} established that $d_{\text{eff}}^{\text{pde}}$ identifies the dimension of the parameter space left unconstrained by the operator equation. We exploit this null space to adapt boundary conditions without disturbing the learned physics.

\subsection{Projection Strategies}
\subsubsection{Parameter Space Projection}
We define the PDE Hessian $H_p$ evaluated at the optimized base parameters $\hat{\bm{\theta}}_B$. This matrix identifies the directions consumed by the PDE constraint. We then construct the orthogonal projection matrix onto the null space of the PDE:

\begin{equation}
    \label{eq:orthogonal_projection}
    \Pi_p = I - H_p H_p^{+}
\end{equation}

The operator $\Pi_p$ removes any component of a parameter update that would alter the PDE residual. To ensure the boundary task cannot interfere with the learned physics, we project the boundary residual Jacobian onto the physics null space:

\begin{equation}
    \label{eq:projecting_boundary_sensitivity}
    V_{\text{FIM}} = \Pi_p \Phi_b
\end{equation}
$V_{\text{FIM}}$ captures the sensitivity of the boundary residuals to parameter changes that lie strictly within the PDE null space. By projecting $\Phi_b$ onto this null space, we isolate the degrees of freedom that can be used to adapt the BCs without affecting the PDE residual. This is crucial for maintaining the integrity of the learned physics while allowing for flexible adaptation to changing BCs. Drawing from orthogonal gradient descent for continual learning \cite{farajtabar2020orthogonal}, the method treats PDE learning and BC satisfaction as discrete sequential objectives, mirroring the sequential approach taken in classical differential equation solving. The PDE is first learned, then the BCs are adapted.

We use a second order Taylor expansion to approximate the model's response to parameter shifts along this basis, allowing us to solve for the optimal coefficients that minimize the boundary residual while respecting the PDE constraints. The Taylor expansion of the model around $\hat{\bm{\theta}}_B$ at boundary points is given by:
\begin{equation}
    g(x_b, \hat{\bm{\theta}}_B+\Delta\bm{\theta}) \approx g(x_b, \hat{\bm{\theta}}_B) + \nabla g(x_b, \hat{\bm{\theta}}_B)^\mathsf{T} \Delta\bm{\theta} + \frac{1}{2} \Delta\bm{\theta}^\mathsf{T} \nabla^2 g(x_b, \hat{\bm{\theta}}_B) \Delta\bm{\theta}
\end{equation}
\noindent where $\nabla g$ and $\nabla^2 g$ are the full-parameter gradient and Hessian of the model output at a boundary point. Since we are updating the parameters within the PDE null space, $\Delta\bm{\theta}$  is constrained to lie within the subspace spanned by the columns of $V_{\text{FIM}}$ ($\Delta\bm{\theta} = \sum_{j=1}^{k} c_j V_{\text{FIM},j} = V_{\text{FIM}}\bm{c}$). We define $g_V$ and $H_V$ as the local gradient and hessian of the model output, mathematically projected onto our retained basis $V_{\text{FIM}} \in \mathbb{R}^{|\bm{\theta}| \times N_{b}}$:

\begin{equation}
    \begin{aligned}
        g_V & = V_{\text{FIM}}^\mathsf{T} \nabla g(x_b, \hat{\bm{\theta}}_B)                 \\
        H_V & = V_{\text{FIM}}^\mathsf{T}\nabla^2 g(x_b, \hat{\bm{\theta}}_B) V_{\text{FIM}}
    \end{aligned}
\end{equation}
Substituting these projections into a Taylor expansion yields a refined model $g_R$ evaluated at the boundary location $x_b$:

\begin{equation}
    g_R(x_b, \hat{\bm{\theta}}_B+\Delta\bm{\theta}) \approx g(x_b, \hat{\bm{\theta}}_B) + g_V^\mathsf{T} \bm{c} + \frac{1}{2} \bm{c}^\mathsf{T} H_V \bm{c}
\end{equation}
The vector $\bm{c} = [c_1, \dots, c_k]^\mathsf{T}$ contains the shift coefficients. Adapting to the new BCs involves solving for these coefficients by minimizing the squared residual between this quadratic approximation and the target boundary value $u_b$. Here, any optimizer can be employed, but we use a Gauss-Newton approach for simplicity. Instead of computing the second-order terms directly, we take multiple smaller first-order steps, re-evaluating the network, its local Jacobian $\Xi(\bm{x}_b) = \nabla_\theta g(\bm{x}_b, \bm{\theta}_{\text{current}})^\mathsf{T} V$ and the null space geometry at each iteration:

\begin{equation}
    \min_{\Delta \bm{c}} \| g(\bm{x}_b, \bm{\theta}_{\text{current}}) + \Xi(\bm{x}_b)\Delta \bm{c} - u_b \|^2 \implies \Delta \bm{c} = \Xi^+(\bm{u}_b - g(\bm{x}_b, \bm{\theta}_{\text{current}}))
\end{equation}

This iterative re-linearization effectively captures the curvature without ever explicitly forming the Hessian matrix. This formulation decouples the task of PDE manifold learning from boundary constraint satisfaction, enabling rapid boundary adaptation. Once the optimal $\bm{c}$ is found, the final parameter update $\Delta\bm{\theta}$ yields the refined state $\bm{\theta}_R = \hat{\bm{\theta}}_B + \Delta\bm{\theta}$. In exact arithmetic, $\Delta\bm{\theta}$ lives entirely within the null space of the $H_p$ and should leave the PDE residual unchanged. In practice, finite precision arithmetic and iterative re-linearization introduce small drifts which we monitor by evaluating the full non-linear PDE loss at $\bm{\theta}_R$. Section~\ref{sec:refinements} addresses this drift via predictor-corrector retraction.

Up to this point, we have outlined a theoretically sound strategy for subspace projection and boundary adaptation in PINNs. However, there are practical challenges that must be addressed to make this approach scalable. In particular, we must address the computational bottlenecks associated with large neural networks and high-dimensional parameter spaces. The explicit projection method requires computing the pseudo-inverse of $H_p$ which is $\mathcal{O}(d^3)$. We switch to a dual formulation that leverages the empirical NTK to improve efficiency.

\subsubsection{Function Space Projection}

The empirical physics Hessian $H_p = \frac{1}{N_p} \boldsymbol{\Phi}_p \boldsymbol{\Phi}_p^T$ shares its null space of the Jacobian $\boldsymbol{\Phi}_p^\mathsf{T}$. Because orthogonal projectors are scale-invariant, the global $1/N_p$ multiplier cancels out during the pseudo-inverse operation, leaving $H_p H_p^+ = \boldsymbol{\Phi}_p \boldsymbol{\Phi}_p^+$. Assuming the interior collocation evaluations yield linearly independent gradients, $\boldsymbol{\Phi}_p$ possesses full column rank. This allows us to explicitly unpack the pseudo-inverse as $\boldsymbol{\Phi}_p\boldsymbol{\Phi}_p^+ = \boldsymbol{\Phi}_p (\boldsymbol{\Phi}_p^T \boldsymbol{\Phi}_p)^{-1} \boldsymbol{\Phi}_p^T$. We define the scaled empirical NTK evaluated across the $N_p$ interior points as $K_{\text{NTK}} = \frac{1}{N_p} \boldsymbol{\Phi}_p^T \boldsymbol{\Phi}_p$. Substituting this into Eq. \eqref{eq:orthogonal_projection} transforms the projector:
\begin{equation}
    \label{eq:ntk_projection}
    \Pi_p = I - \frac{1}{N_p} \boldsymbol{\Phi}_p K_{\text{NTK}}^{-1} \boldsymbol{\Phi}_p^T
\end{equation}
Substituting Eq.~\eqref{eq:ntk_projection} into Eq.~\eqref{eq:projecting_boundary_sensitivity} gives the boundary sensitivity matrix in dual form:
\begin{equation}
    \begin{aligned}
        V_{\text{NTK}} & = \left( I - \frac{1}{N_p} \boldsymbol{\Phi}_p K_{\text{NTK}}^{-1} \boldsymbol{\Phi}_p^T \right) \boldsymbol{\Phi}_b      \\
                       & = \boldsymbol{\Phi}_b - \frac{1}{N_p} \boldsymbol{\Phi}_p K_{\text{NTK}}^{-1} (\boldsymbol{\Phi}_p^T \boldsymbol{\Phi}_b) \\
    \end{aligned}
\end{equation}
The final parameter update $\Delta \theta = V_{\text{NTK}} (V_{\text{NTK}}^T V_{\text{NTK}})^{-1}(\bm{u}_b - g(\bm{x}_b, \bm{\theta}_{\text{current}}))$ functions exactly as it did in the parameter-space formulation. The difference is purely computational. We bypass computing the intractable $d \times d$ pseudo-inverse required for $\Pi_p$ and invert the $N_p \times N_p$ NTK matrix instead. In the overparameterized regime ($N_p \ll d$), this dual formulation drops the bottleneck to a fraction of its original cost. The formulation scales to networks with $d\gg N_p$ parameters, provided $N_p$ remains small enough to invert $K_{\text{NTK}}$ efficiently.

\subsection{1D Boundary Adaptation (Steady State Heat Equation)}

We evaluate the proposed framework on a 1D steady-state heat equation~\eqref{eq:ode2}, using the null space identified in Section \ref{sec:invariance} to adapt boundary conditions. The initial training phase focuses exclusively on the PDE constraint. The architecture is a fully-connected feedforward neural network utilizing a single hidden layer of 100 neurons paired with hyperbolic tangent activation functions. We use 100 collocation points, distributed with equal spacing across the domain, alongside two boundary points at $x=-1$ and $x=1$. Optimization follows a two-stage process. We begin with full-batch training for 10,000 epochs via the Adam optimizer (learning rate of $1 \times 10^{-3}$ and exponential decay with $\gamma=0.9998$) and conclude with up to 37,500 epochs using L-BFGS.

\begin{figure}[ht!]
    \centering
    \includegraphics[width=0.95\textwidth]{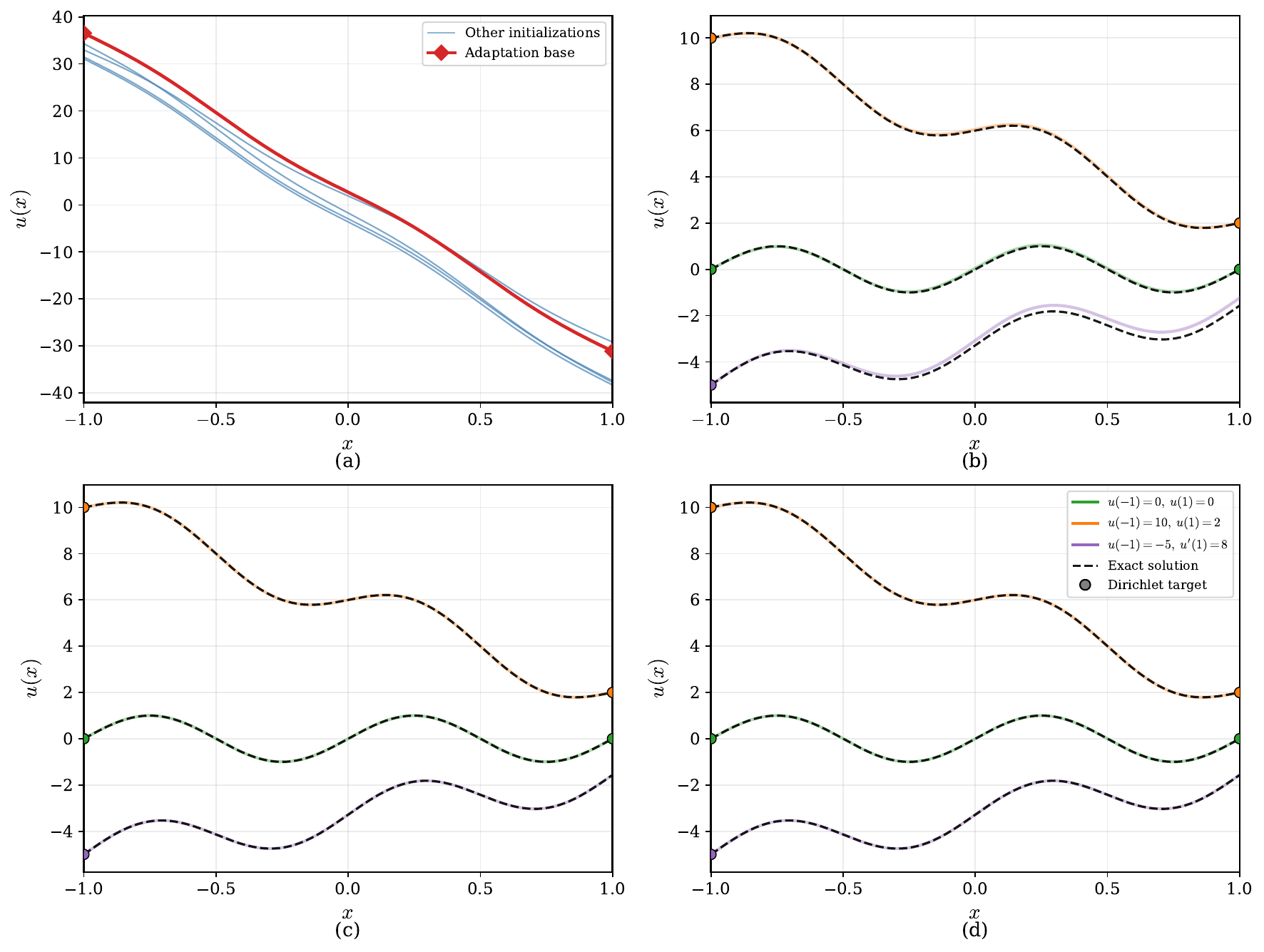}
    \caption{Boundary adaptation in 1D. (a) Multiple PINNs trained exclusively on the PDE residual with different random seeds, each converging to a distinct member of the physics-satisfying solution manifold. The red curve is the adaptation base used in (b)--(d). (b) Full-network subspace projection without a predictor-corrector step. (c) Last-layer subspace projection. (d) Full-network subspace projection with the predictor-corrector step. Colored solid curves are adapted PINN solutions}
    \label{fig:sine_subspace_projection_results}
\end{figure}

As Figure \ref{fig:sine_subspace_projection_results}(a) illustrates, the resulting solution is sensitive to initialization. Different initialization weights yield distinct functions that all satisfy the governing PDE. To adapt the BCs, we perform the subspace projection within the null space of the PDE constraint. Our first adaptation approach utilized the full network's physics null space to adapt to the BCs. We tested three specific scenarios: a symmetric shift ($u(-1) = 0, u(1) = 0$), an asymmetric shift ($u(-1) = 10, u(1) = 2$), and a mixed BC case ($u(-1) = -5, u'(1) = 8$).

Performance metrics reveal a trade-off, with the mixed BC case proving more challenging than the symmetric and asymmetric Dirichlet cases (Table~\ref{tab:1d_adaptation_summary} and Figure \ref{fig:sine_subspace_projection_results}(b)). As shown in Table~\ref{tab:1d_adaptation_summary}, the PDE residual climbed from the base value of $2.16 \times 10^{-6}$, a drift that is a logical consequence of the iterative Gauss-Newton method and the rcond threshold. The neural network manifold is inherently curved. We navigate this surface by taking small steps along the tangent plane, so error inevitably accumulates. For these specific adaptations, we moved from the base PDE solution to the final BC state in 100 steps. While increasing the step count would likely sharpen the adaptation and reduce residual growth, it would do so at the expense of computational speed. In addition, due to finite precision computation, the $10^{-10}$ rcond threshold might mistakenly push some physics directions into the null space. Lowering rcond can mitigate this issue at the risk of pushing actual null space directions into the physics space. Even with these difficulties, the BC adaptation remains effective, shifting the boundary values with controlled error.

\begin{table}[htbp!]
    \centering
    \caption{Summary of 1D boundary adaptation performance across methods and BC scenarios.}
    \label{tab:1d_adaptation_summary}
    \resizebox{0.8\textwidth}{!}{
        \begin{tabular}{llcc}
            \toprule
            \textbf{Method} & \textbf{BC Case} & \textbf{RMSE}         & \textbf{PDE Residual} \\
            \midrule
            \multirow{3}{*}{\shortstack[l]{Last-layer}}
                            & Symmetric        & $4.79 \times 10^{-6}$ & $4.10 \times 10^{-6}$ \\
                            & Asymmetric       & $4.30 \times 10^{-6}$ & $3.93 \times 10^{-6}$ \\
                            & Mixed            & $7.80 \times 10^{-5}$ & $4.17 \times 10^{-6}$ \\
            \midrule
            \multirow{3}{*}{\shortstack[l]{Full proj.                                          \\(predictor-only)}}
                            & Symmetric        & $3.38 \times 10^{-2}$ & $9.81 \times 10^{-2}$ \\
                            & Asymmetric       & $3.21 \times 10^{-2}$ & $7.06 \times 10^{-2}$ \\
                            & Mixed            & $2.19 \times 10^{-1}$ & $1.13 \times 10^{-1}$ \\
            \midrule
            \multirow{3}{*}{\shortstack[l]{Full proj.                                          \\(pred.-corrector)}}
                            & Symmetric        & $2.69 \times 10^{-4}$ & $4.52 \times 10^{-6}$ \\
                            & Asymmetric       & $1.80 \times 10^{-4}$ & $4.47 \times 10^{-6}$ \\
                            & Mixed            & $4.00 \times 10^{-4}$ & $6.35 \times 10^{-6}$ \\
            \midrule
            \multirow{3}{*}{\shortstack[l]{Last-layer                                          \\(regime anchoring)}}
                            & Symmetric        & $2.12 \times 10^{-6}$ & $2.51 \times 10^{-6}$ \\
                            & Asymmetric       & $2.12 \times 10^{-6}$ & $2.50 \times 10^{-6}$ \\
                            & Mixed            & $5.38 \times 10^{-5}$ & $2.52 \times 10^{-6}$ \\
            \midrule
            \multirow{3}{*}{\shortstack[l]{Full proj.                                          \\(regime anch., pred.-only)}}
                            & Symmetric        & $6.25 \times 10^{-6}$ & $2.52 \times 10^{-6}$ \\
                            & Asymmetric       & $1.45 \times 10^{-3}$ & $4.07 \times 10^{-5}$ \\
                            & Mixed            & $3.44 \times 10^{-4}$ & $5.89 \times 10^{-6}$ \\
            \midrule
            \multirow{3}{*}{\shortstack[l]{Full proj.                                          \\(regime anch., pred.-corr.)}}
                            & Symmetric        & $8.90 \times 10^{-8}$ & $1.06 \times 10^{-8}$ \\
                            & Asymmetric       & $8.60 \times 10^{-6}$ & $5.69 \times 10^{-9}$ \\
                            & Mixed            & $6.85 \times 10^{-6}$ & $1.48 \times 10^{-8}$ \\
            \bottomrule
        \end{tabular}
    }
\end{table}

We now examine layer-wise behavior to identify which layers anchor the boundary conditions. We begin by calculating $d_{\text{eff}}$ for each individual layer. For this single hidden layer architecture, the effective dimension of the physics constraint is not localized. It is distributed uniformly across the entire architecture. With $d_{\text{eff}}^{\text{pde}}$ holding at 2.00 for every layer and the network as a whole, it is clear that each layer maintains unconstrained directions that permit boundary adaptation. In contrast, the BCs exhibit a layer preference. We found that $d_{\text{eff}}^{\text{bc}}$ was lower in the last layer (12.00) and higher in the first layer (16.00). Since layers differ in parameter count and therefore baseline capacity, the meaningful comparison is the ratio $d_{\text{eff}}^{\text{bc}}/d_{\text{eff}}^{\text{pde}}$, which normalizes BC consumption against the layer's PDE-surviving capacity. By this metric the last layer has the lowest ratio, indicating that the network primarily anchors the BCs within its final layer. This insight allows us to move beyond full-network updates toward layer-wise projection. By keeping other layers fixed and adapting only the final layer, we can exploit the fact that this layer is linear with respect to its parameters, reducing the adaptation to a single projected least-squares solve. This targeted approach yielded superior results across all three scenarios (Table~\ref{tab:1d_adaptation_summary} and Figure \ref{fig:sine_subspace_projection_results}(c)), improving both RMSE and PDE residuals over the full projection method. These findings show that for this problem final-layer adaptation offers a more precise fit for new BCs while preserving the integrity of the learned physics. Furthermore, optimizing fewer parameters reduces computational overhead. Existing transfer learning methods leveraged this layer-wise behavior empirically~\cite{desai2022one,goswami2020transfer}. $d_{\text{eff}}$ provides a geometric explanation for why it works. For this 1D problem, $d_{\text{eff}}$ identifies the final layer as the most BC-consumed. We investigate whether this layer-wise heuristic generalizes to higher dimensions in Section~\ref{sec:multi_dimensional}.

\section{Refinements for Residual Fidelity}
\label{sec:refinements}

\subsection{Predictor-Corrector Riemannian Retraction}

PDE residual drift during boundary adaptation is problematic because derived quantities such as heat flux inherit the residual error. A few-order-of-magnitude jump in the residual translates to comparable error in any quantities extracted from the network's derivatives. We have already pointed to the partial mitigation strategy of increasing the number of steps to navigate the subspace more tightly and the more informed strategy of finding where the BCs are primarily anchored. In this section, we propose a predictor-corrector Riemannian retraction~\cite{absil2008optimization} to handle the case where the BCs are not primarily anchored in the last layer.

The predictor step executes the null space boundary adaptation. The corrector step performs a strictly orthogonal adaptation in the range space of the physics operator. The total parameter update decomposes into two orthogonal components:

\begin{equation}
    \begin{aligned}
        \Delta \bm{\theta}       & = \Delta \bm{\theta}_{bc} + \Delta \bm{\theta}_{pde}            \\
        \Delta \bm{\theta}_{bc}  & = V_{\text{FIM}} \Delta \bm{c}_{bc}                             \\
        \Delta \bm{\theta}_{pde} & = \bm{\Phi}_{p}^+\bm{R}(\bm{x}_p, \bm{\theta}_{\text{current}})
    \end{aligned}
\end{equation}

\noindent where $\Delta \bm{\theta}_{bc}$ represents the predictor step, which slides the network along the PDE null space to satisfy the new BCs. $\Delta \bm{\theta}_{pde}$ is the corrector step, which drives any accumulated PDE residual back down to zero by moving orthogonally to the PDE null space. $\bm{R}=f(\bm{x}_p)-\mathcal{L}g(\bm{x}_p,\bm{\theta}_{\text{current}})$ at all collocation points. The predictor step is identical to our original subspace projection method.

With the predictor-corrector retraction, the PDE residual stays at the same order of magnitude as the base model after adaptation (Table~\ref{tab:1d_adaptation_summary}), while the RMSE on the new BCs matches or improves on the predictor-only result (Figure~\ref{fig:sine_subspace_projection_results}(d)). The predictor-corrector retraction extends the applicability of subspace projection to BCs that are far from the base model training conditions, where predictor steps accumulate substantial drift. For the last layer adaptation, the corrector is not needed because the space is already flat. The corrector is only needed if we are interested in adapting deep layers. The corrector step introduces negligible additional computational overhead: $\bm{\Phi}_p^+\bm{R}$ requires only a matrix-vector product against $\bm{\Phi}_p^+$, which is already cached from the predictor step.

Predictor-only and predictor-corrector adaptation both take approximately 1.5--2.0 seconds while the last-layer adaptation takes under 0.02 seconds. The corrector adds no measurable overhead, confirming the cache-reuse argument. While we use ordinary least squares for the corrector solve, practitioners with alternative preferences can substitute any linear solver without affecting the geometric structure.

\subsection{Regime Anchoring for Adaptation}

\begin{figure}[ht!]
    \centering
    \includegraphics[width=0.95\textwidth]{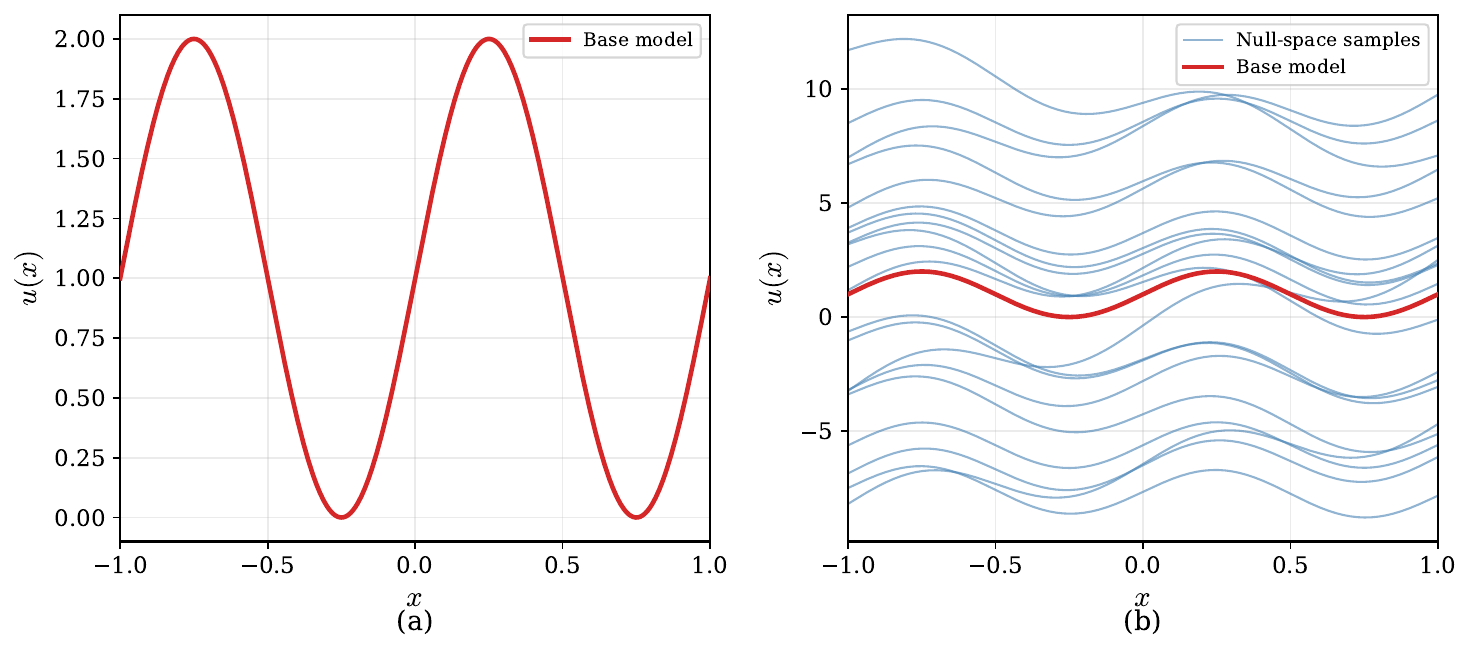}
    \caption{Regime anchoring and last-layer null-space variability. (a) Base PINN trained with a small BC loss ($\lambda_b = 10^{-3}$, $\lambda_p = 1$) anchoring the solution near $u=1$ at both boundaries. (b) Random directions drawn from the last-layer physics null space. The family of curves spans a wide range of boundary values with preserved physics residual.}
    \label{fig:sine_subspace_projection_results_kinda_close_bc}
\end{figure}

When the approximate range of expected boundary conditions is known a priori, the base training can be biased toward this regime. We refer to this as regime anchoring: the base model satisfies the PDE strictly and approximately satisfies a representative set of boundary values, so that subspace projection later refines the BCs from a nearby starting point rather than from an arbitrary member of the physics-satisfying manifold. Any standard PINN weighting scheme can provide this bias --- NTK-based~\cite{wang2022and}, self-adaptive~\cite{mcclenny2023self}, or manually tuned weights.

Regime anchoring is especially important for homogeneous problems where the trivial solution $u \equiv 0$ is a strong attractor. Pre-training on PDE alone risks collapse to the trivial solution, and the BC anchor steers the optimizer toward physically relevant solutions.

As shown in Figure~\ref{fig:sine_subspace_projection_results}(a), the base PINN without anchoring converges far from the target BCs, which causes PDE residual drift during adaptation in the absence of a corrector step. Introducing a small BC loss during pre-training ($\lambda_b = 10^{-3}$, $\lambda_p = 1$) draws the base solution toward the anticipated BC range, shortening the null-space path to the target. We tested this with anticipated BCs near $u = 1$ at both boundaries (Figure~\ref{fig:sine_subspace_projection_results_kinda_close_bc}(a)). Across both last-layer and full projection methods, regime anchoring produced the lowest RMSE and PDE residual (Table~\ref{tab:1d_adaptation_summary}), because the null-space path from the anchored base to the target BC is shorter than from an arbitrary kernel member. Regime anchoring's adaptation takes approximately 1.6 seconds, slightly faster than the unanchored full-projection methods due to the shorter null-space path.

\subsection{Sample Generation in the Physics Null Space}

Beyond boundary adaptation, the null-space projector enables a distinct application: sampling parameter configurations that share the base model's PDE residual but produce varying boundary values. The orthogonal projector $\Pi_p$~\eqref{eq:orthogonal_projection} annihilates every component of a parameter perturbation that would alter the PDE residual. Drawing $\boldsymbol{z} \sim \mathcal{N}(\mathbf{0}, \mathbf{I})$ and forming $\boldsymbol{\theta}_s = \hat{\boldsymbol{\theta}}_B + \Pi_p \boldsymbol{z}$ yields a parameter configuration whose PDE residual is identical to the base model while the boundary values shift freely. Figure~\ref{fig:sine_subspace_projection_results_kinda_close_bc}(b) visualizes a family of such samples.

The 1D experiments establish that subspace projection can adapt boundary conditions post-hoc using either the full network or the final layer alone, with the final layer preferred when
computationally feasible. Whether this final-layer anchoring extends
to higher dimensions is the subject of Section~\ref{sec:multi_dimensional}.

\section{Multi-Dimensional Scaling and Extent of Layer-Wise Capacity}
\label{sec:multi_dimensional}

The 1D experiments showed that subspace projection adapts boundary conditions effectively, and that the final layer is the most BC-consumed for shallow tanh networks. This section tests whether these findings extend to multi-dimensional problems with richer architectures. We use 2D Poisson as a representative elliptic operator and a 3-hidden-layer residual network as a representative deeper architecture. Here, we employ rcond of $10^{-8}$. We then study the 1D Burgers' equation as a representative non-linear evolution operator.

\subsection{2D Poisson equation}
\label{sec:2d_poisson}
The governing equation is defined as:

\begin{equation}
    \label{eq:poisson_equation}
    \Delta u = -2\pi^2\sin(\pi x)\sin(\pi y), \quad (x, y) \in [-1, 1] \times [-1, 1]
\end{equation}

We use a residual network architecture comprising 3 hidden layers with 100 neurons each. To evaluate the impact of nonlinearity, we train a tanh network using hyperbolic tangent activations and a SiLU network using SiLU (Swish) activations. We use a $15 \times 15$ uniform grid of collocation points (225 total) with no boundary conditions applied during base training. After training, we apply subspace projection with predictor-corrector to adapt the model to the BCs of the problem: $u(-1, y) = u(1, y) = u(x, -1) = 0$ and $u(x, 1) = \sin(\pi x)$. The exact solution is $u^*(x, y) = \sin(\pi x)\sin(\pi y)$.

\begin{figure}[ht!]
    \centering
    \includegraphics[width=0.95\textwidth]{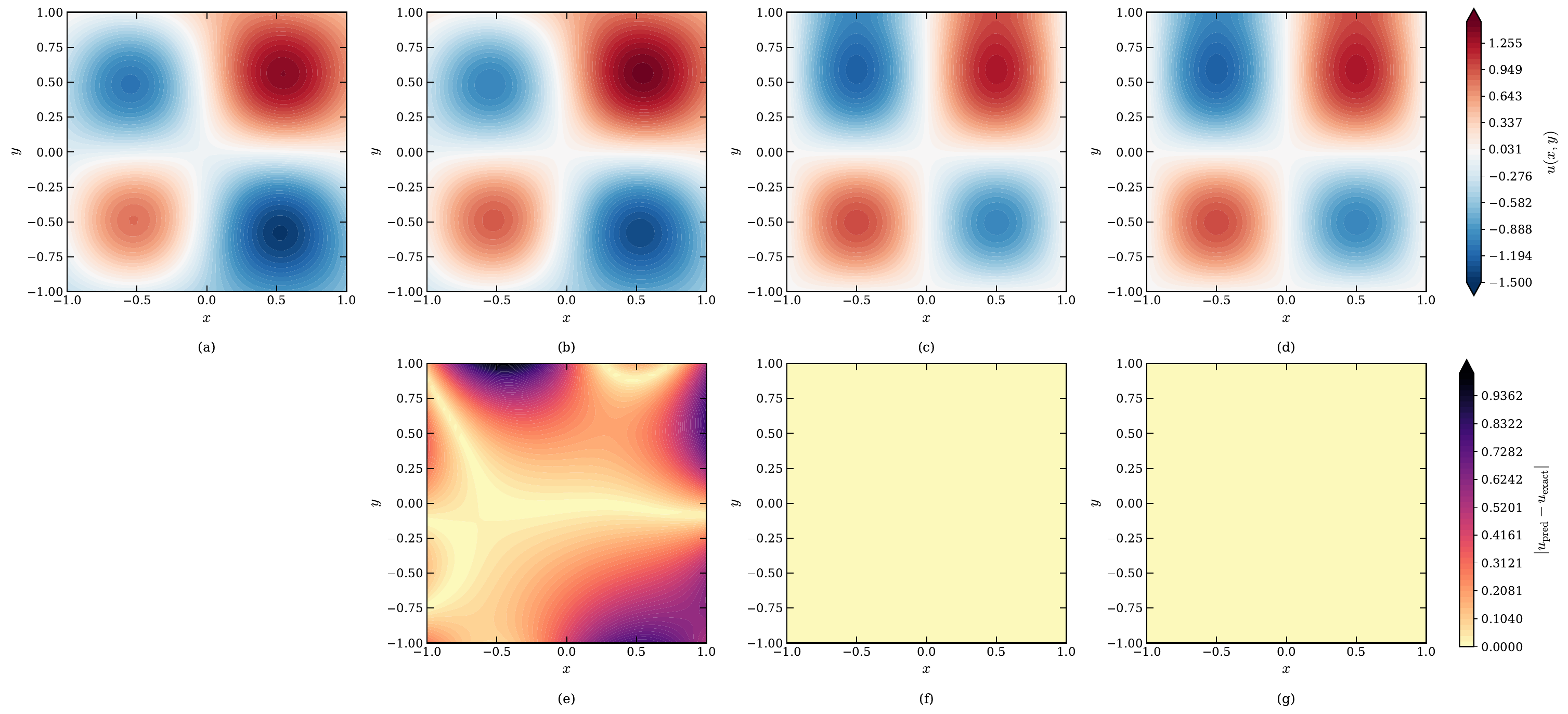}
    \caption{Tanh network. (a) Baseline PINN trained exclusively on the PDE residual. (b) Boundary adaptation restricted to Layer $3$ (output), resulting in the significant structured error observed in (e). (c) Adaptation using only Layer $2$ (penultimate), with the corresponding error map in (f). (d) Successful adaptation achieved by expanding the projection to Layers $3$ and $2$. The pooling of surviving parameter directions restores the solution to near-ground truth, as evidenced by the negligible residual error in (g).}
    \label{fig:tanh_poisson_subspace_projection_results}
\end{figure}

\begin{figure}[ht!]
    \centering
    \includegraphics[width=0.95\textwidth]{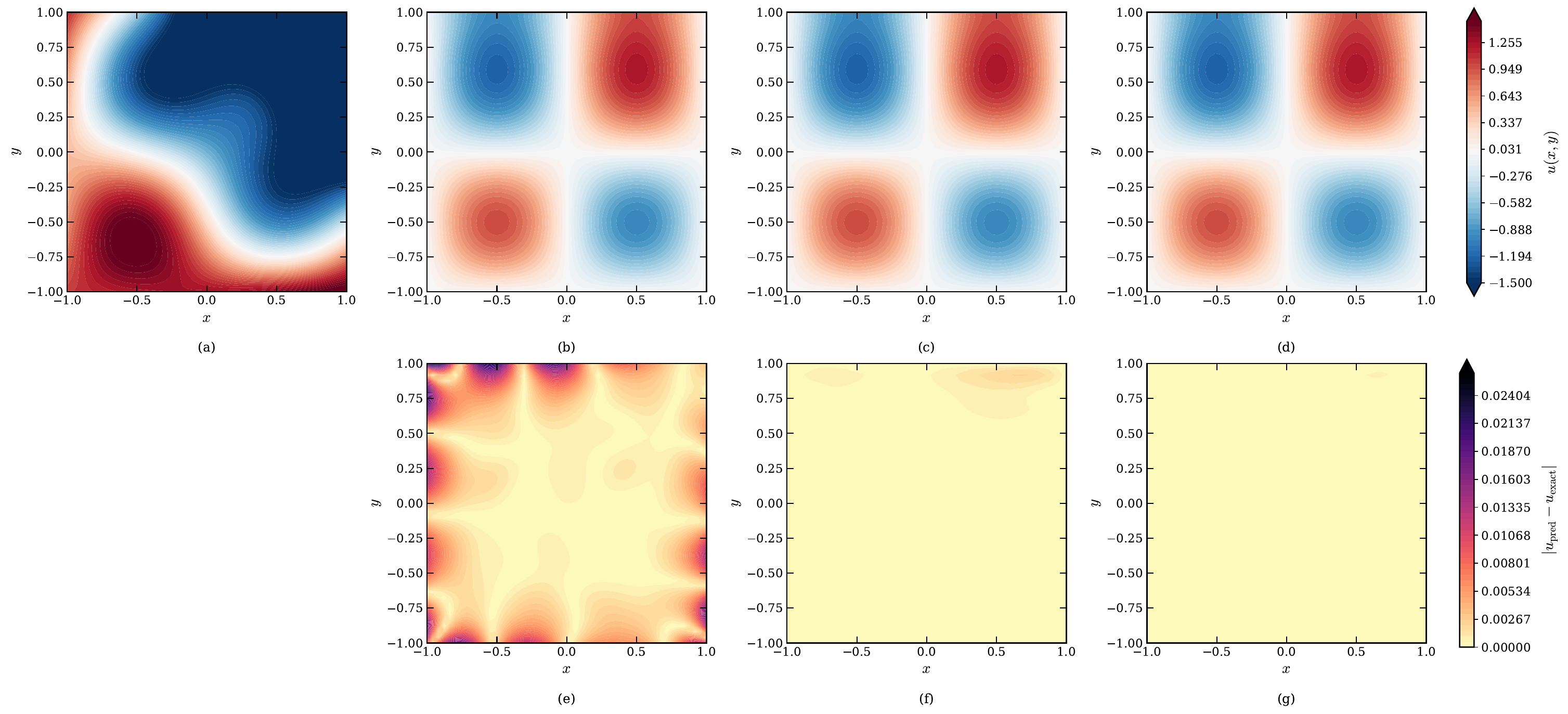}
    \caption{SiLU network. (a) Baseline PINN trained on PDE residuals. (b) Solution adapted using only Layer $3$ (output), with corresponding error in (e). (c) Solution adapted using only Layer $2$ (penultimate), with corresponding error in (f). (d) Solution adapted using Layers $3$ and $2$, with corresponding error in (g).}
    \label{fig:silu_poisson_subspace_projection_results}
\end{figure}

Figures~\ref{fig:tanh_poisson_subspace_projection_results} and \ref{fig:silu_poisson_subspace_projection_results} show the results for the tanh and SiLU networks and Table~\ref{tab:poisson_adaptation} presents the associated errors after BC adaptation. The base training phases yield visibly distinct solutions (Figures~\ref{fig:tanh_poisson_subspace_projection_results}(a) and \ref{fig:silu_poisson_subspace_projection_results}(a)). Single-layer adaptation on the final layer fails to satisfy the BCs for both architectures (Figures~\ref{fig:tanh_poisson_subspace_projection_results}(b,e), \ref{fig:silu_poisson_subspace_projection_results}(b,e)). The tanh network's terminal layer yields RMSE $3.16 \times 10^{-1}$, a clear failure. The SiLU terminal layer achieves lower RMSE ($4.36 \times 10^{-3}$) but at the cost of severe PDE degradation ($3.95 \times 10^{-3}$, four orders of magnitude above the base). The last-layer null space lacks sufficient capacity to enforce the new BCs without corrupting the learned physics.

The qualitative finding is that adapting only the final layer does not extend to 2D Poisson. Even when RMSE improves (SiLU terminal), the single-layer null space cannot simultaneously enforce the new BCs and preserve the learned physics. Expanding the adaptation to the final two layers restores accurate adaptation for both architectures (Figures~\ref{fig:tanh_poisson_subspace_projection_results}(d,g), \ref{fig:silu_poisson_subspace_projection_results}(d,g), Table~\ref{tab:poisson_adaptation}). This indicates that the adaptation scope must scale with problem dimensionality.

\begin{table}[htbp!]
    \centering
    \caption{2D Poisson boundary adaptation performance by architecture and adaptation scope. Base PDE residual is $\mathcal{O}(10^{-7})$ for both architectures.}
    \label{tab:poisson_adaptation}
    \begin{tabular}{llcc}
        \toprule
        \textbf{Layers Adapted} & \textbf{Architecture} & \textbf{RMSE}         & \textbf{PDE Residual} \\
        \midrule
        \multirow{2}{*}{Last-layer}
                                & Tanh                  & $3.16 \times 10^{-1}$ & $8.08 \times 10^{-8}$ \\
                                & SiLU                  & $4.36 \times 10^{-3}$ & $3.95 \times 10^{-3}$ \\
        \midrule
        \multirow{2}{*}{Penultimate-layer}
                                & Tanh                  & $3.21 \times 10^{-3}$ & $3.61 \times 10^{-7}$ \\
                                & SiLU                  & $3.12 \times 10^{-4}$ & $7.57 \times 10^{-7}$ \\
        \midrule
        \multirow{2}{*}{Two-layer}
                                & Tanh                  & $1.25 \times 10^{-3}$ & $1.45 \times 10^{-8}$ \\
                                & SiLU                  & $9.76 \times 10^{-5}$ & $1.54 \times 10^{-6}$ \\
        \bottomrule
    \end{tabular}
\end{table}

The 2D Poisson operator fixes what $d_{\text{eff}}^{\text{pde}}$
measures. Poisson is second order, so any metric that merely counted derivative
order would return a small finite integer, exactly as the second-order ODE did
in Section~\ref{sec:invariance}. Instead $d_{\text{eff}}^{\text{pde}}$ does not
saturate to an integer. It increases with network width and depth
(Table~\ref{tab:capacity_ratio_prediction}). This is the expected behavior if
the metric tracks the dimension of the operator's kernel rather than its order:
the Laplacian's kernel is the infinite-dimensional space of harmonic functions,
and a finite network can represent only a capacity-limited slice of it. The
low-order ODEs and the Poisson operator therefore make different predictions
under the two readings, and the measured behavior falls on the kernel-dimension
side. For the ODEs, kernel dimension and operator order coincide, but Poisson is the
case that separates them. The values in Table~\ref{tab:capacity_ratio_prediction}
should accordingly be read as the network's finite-dimensional representational
bandwidth for the harmonic kernel, not as a recovered analytical invariant.

The same fact shapes how the metric is used for layer selection. In the 1D setting, $d_{\text{eff}}^{\text{pde}}$ saturates to the kernel dimension independent of architecture, so layer-to-layer variation in the ratio $d_{\text{eff}}^{\text{bc}}/d_{\text{eff}}^{\text{pde}}$ is driven entirely by $d_{\text{eff}}^{\text{bc}}$, which is a clean capacity signal. In the 2D Poisson setting, $d_{\text{eff}}^{\text{pde}}$ varies across layers because each layer represents a different finite-dimensional approximation of the harmonic kernel. A layer with a larger $d_{\text{eff}}^{\text{pde}}$ may simply have higher raw capacity rather than being structurally less constrained, so the ratio can conflate BC consumption with baseline layer capacity. Whether this conflation matters in practice is problem-dependent, and the results below show that the ratio provides a useful but approximate guide whose signal strength varies with the activation function.

\begin{table}[htbp!]
    \centering
    \caption{Predicting the optimal single-layer adaptation target using the $d_{\text{eff}}^{\text{bc}} / d_{\text{eff}}^{\text{pde}}$ ratio heuristic for Tanh and SiLU networks.}
    \label{tab:capacity_ratio_prediction}
    \renewcommand{\arraystretch}{1.2}
    \resizebox{\textwidth}{!}{
        \begin{tabular}{l l c c c c c}
            \toprule
            \textbf{Architecture} & \textbf{Candidate Layer} & $\mathbf{d_{\text{eff}}^{\text{pde}}}$ & $\mathbf{d_{\text{eff}}^{\text{bc}}}$ & \textbf{Ratio} & \textbf{RMSE}                  & \textbf{PDE Residual} \\
            \midrule
            \multirow{4}{*}{Tanh}
                                  & Layer $3$ (output)       & 4.239                                  & 49.848                                & 11.761         & $3.16 \times 10^{-1}$          & $8.08 \times 10^{-8}$ \\
                                  & Layer $2$ (penultimate)  & 25.000                                 & 105.348                               & \textbf{4.214} & $\mathbf{3.21 \times 10^{-3}}$ & $3.61 \times 10^{-7}$ \\
                                  & Layer $1$                & 18.740                                 & 86.708                                & 4.627          & $1.88 \times 10^{-3}$          & $7.54 \times 10^{-5}$ \\
                                  & Layer $0$ (input)        & 12.173                                 & 65.813                                & 5.406          & $2.01 \times 10^{0}$           & $8.49 \times 10^{1}$  \\
            \midrule
            \multirow{4}{*}{SiLU}
                                  & Layer $3$ (output)       & 10.930                                 & 25.298                                & \textbf{2.315} & $\mathbf{4.36 \times 10^{-3}}$ & $3.95 \times 10^{-3}$ \\
                                  & Layer $2$ (penultimate)  & 19.532                                 & 47.516                                & 2.433          & $3.12 \times 10^{-4}$          & $7.57 \times 10^{-7}$ \\
                                  & Layer $1$                & 16.809                                 & 41.495                                & 2.469          & $9.46 \times 10^{-4}$          & $1.37 \times 10^{-6}$ \\
                                  & Layer $0$ (input)        & 13.532                                 & 33.900                                & 2.505          & $6.67 \times 10^{-2}$          & $1.12 \times 10^{0}$  \\
            \bottomrule
        \end{tabular}
    }
\end{table}

Table~\ref{tab:capacity_ratio_prediction} shows per-layer adaptation performance for both architectures. For the Tanh network, the ratio heuristic identifies Layer $2$ as the primary target (lowest ratio $4.214$, RMSE $3.21 \times 10^{-3}$), which is near-optimal. Layer $1$ achieves a slightly lower RMSE ($1.88 \times 10^{-3}$) at a higher ratio ($4.627$). For the SiLU network, the heuristic nominally identifies Layer $3$ (lowest ratio $2.315$), but Layer $2$ yields a substantially lower RMSE ($3.12 \times 10^{-4}$ versus $4.36 \times 10^{-3}$). In the SiLU case the ratios are tightly clustered ($2.315$--$2.505$), so the heuristic provides a weaker signal than in the Tanh case, where the lowest ratio ($4.214$) is more clearly separated from the remaining layers. The ratio therefore serves as a useful but approximate guide for layer selection.

\subsection{Subspace Projection vs.\ Gradient-Based Adaptation}
\label{sec:comparison}

We compare three adaptation strategies applied to the same physics-trained SiLU ResNet from Section~\ref{sec:2d_poisson}. All three methods start from an identical base model ($\text{PDE MSE} = 3.35 \times 10^{-7}$, $\text{BC MSE} = 6.67$). Subspace projection is implemented with 100 incremental steps and the predictor-corrector retraction, completing in $t_{\text{sub}} = 6.0\,\text{s}$. Both full fine-tuning and last-layer fine-tuning run for 10{,}000 Adam steps followed by 30{,}000 L-BFGS steps. Loss weights for both the PDE and BC objectives are balanced using the adaptive scheme of~\cite{wang2021understanding} and the L-BFGS phase inherits the final weights from Adam.

\begin{table}[htbp!]
    \centering
    \caption{Comparison of boundary adaptation strategies on the 2D Poisson problem. BC MSE is evaluated against the target boundary values. Fine-tuning results are reported both at $t_{\text{sub}} = 6.0\,\text{s}$ (equal-time) and at training end.}
    \label{tab:comparison}
    \begin{tabular}{lcccc}
        \toprule
        \textbf{Method}                              & \textbf{Time (s)} & \textbf{PDE MSE}    & \textbf{BC MSE}     & \textbf{RMSE}       \\
        \midrule
        Base model                                   & ---               & $3.35\times10^{-7}$ & $6.67\times10^{0}$  & ---                 \\
        Subspace projection                          & $6.0$             & $3.13\times10^{-5}$ & $2.16\times10^{-8}$ & $3.78\times10^{-4}$ \\
        \midrule
        Full fine-tuning (at $t_{\text{sub}}$)       & $6.0$             & $1.92\times10^{-4}$ & $2.79\times10^{0}$  & $5.31\times10^{-1}$ \\
        Full fine-tuning (at end)                    & $940.9$           & $4.45\times10^{-7}$ & $2.64\times10^{-9}$ & $3.63\times10^{-5}$ \\
        \midrule
        Last-layer fine-tuning (at $t_{\text{sub}}$) & $6.0$             & $1.02\times10^{-5}$ & $5.51\times10^{0}$  & $1.55\times10^{0}$  \\
        Last-layer fine-tuning (at end)              & $969.5$           & $2.09\times10^{-3}$ & $2.59\times10^{-5}$ & $2.07\times10^{-3}$ \\
        \bottomrule
    \end{tabular}
\end{table}

\begin{figure}[ht!]
    \centering
    \includegraphics[width=0.95\textwidth]{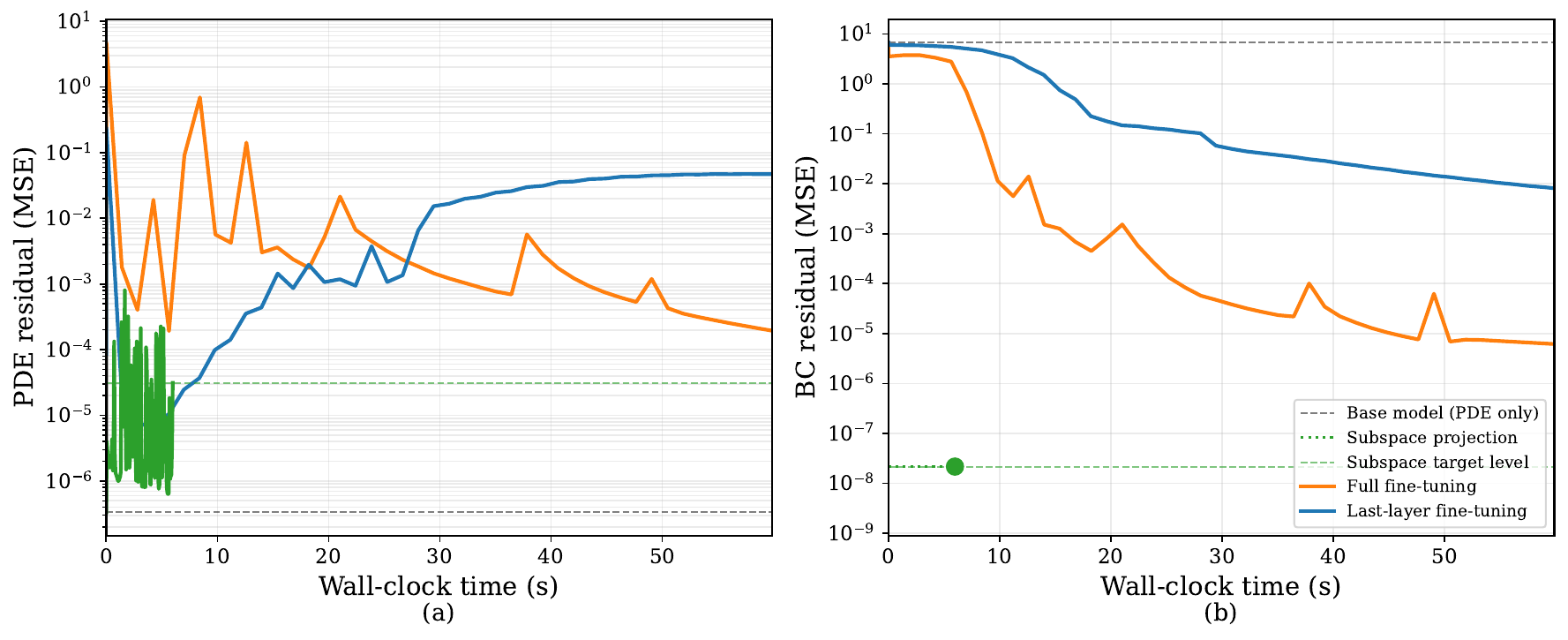}
    \caption{(a) Trajectory of PDE MSE and (b) BC MSE over wall-clock time for three adaptation strategies. The subspace projection result is shown as a single dot at $t_{\text{sub}} = 6.0\,\text{s}$ since intermediate BC residuals measure against the full target throughout the incremental process and are not directly comparable to the fine-tuning trajectories. All methods start from the same base model (dashed line).}
    \label{fig:poisson_comparison}
\end{figure}

The results are summarized in Table~\ref{tab:comparison} and Figure~\ref{fig:poisson_comparison}. Subspace projection completes in $t_{\text{sub}} = 6.0\,\text{s}$, enforcing the new boundary values to a BC MSE of $2.16 \times 10^{-8}$ and achieving RMSE $3.78 \times 10^{-4}$. Despite operating in the physics null space, a transient PDE MSE spike is observed during adaptation — a finite-precision artifact discussed in Section~\ref{sec:invariance} — and the predictor-corrector retraction brings it back to $3.13 \times 10^{-5}$. Within the same $6.0\,\text{s}$ budget, gradient fine-tuning has not yet converged: full fine-tuning yields RMSE $5.31 \times 10^{-1}$ with PDE MSE degraded to $1.92 \times 10^{-4}$, while last-layer fine-tuning preserves the PDE MSE ($1.02 \times 10^{-5}$) but fails to move the BC MSE ($5.51$, near the base value of $6.67$), yielding RMSE $1.55$.

In this problem, gradient-based fine-tuning does not suffer a solution-quality failure at convergence: at $940.9\,\text{s}$, full fine-tuning achieves RMSE $3.63\times10^{-5}$ and BC MSE $2.64\times10^{-9}$, substantially better than subspace projection on both metrics. The honest characterization of this result is that subspace projection delivers near-equivalent practical utility in $6.0\,\text{s}$ — $157\times$ less wall-clock time — with minimal tuning overhead, while full fine-tuning eventually wins on final accuracy. Full fine-tuning reaches the subspace BC target ($2.16 \times 10^{-8}$) at $t = 205.5\,\text{s}$, which is $34.2\times t_{\text{sub}}$. At that crossing the PDE MSE is $2.85 \times 10^{-6}$, two orders of magnitude above the subspace projection value, before eventually recovering. Last-layer fine-tuning never reaches the target BC accuracy; its best BC MSE is $2.50 \times 10^{-5}$ at $t = 243.6\,\text{s}$, and the method converges to RMSE $2.07 \times 10^{-3}$ with a degraded PDE MSE of $2.09 \times 10^{-3}$. This can be observed in Figure~\ref{fig:poisson_comparison} where the PDE MSE rises while the BC MSE falls in the longer run.

\subsection{Nonlinear Operator: 1D Burgers' Equation}
\label{sec:burgers}

The preceding subsections applied subspace projection to a linear elliptic operator. To test the framework under genuinely nonlinear dynamics, we apply it to the viscous Burgers' equation:
\begin{equation}
    \label{eq:burgers}
    \frac{\partial u}{\partial t} + u\frac{\partial u}{\partial x} = \nu \frac{\partial^2 u}{\partial x^2}, \quad (x,t) \in [-1,1]\times[0,1]
\end{equation}
with viscosity $\nu = 0.01/\pi$. At this viscosity, the nonlinear convective term $uu_x$ drives the formation of a steep shock front near $x=0$. The shock location, width, and strength depend sensitively on the initial and boundary data, making this a demanding test for any adaptation strategy.

The base model is trained on the initial condition $u(x,0)=-\sin(\pi x)$ with homogeneous Dirichlet boundaries $u(\pm 1,t)=0$, following the regime-anchoring strategy of Section~\ref{sec:refinements} with manually specified weights ($\lambda_p=\lambda_b=1$). The architecture is a 4-hidden-layer MLP with 120 neurons per layer and $\tanh$ activations. Training uses Adam optimizer followed by an L-BFGS polish, with the PDE discretized on a $200\times160=32{,}000$-point collocation grid. The trained base model reaches a PDE MSE of $1.18\times10^{-6}$. Adaptations compare against a finite-difference reference computed via upwind differencing with BDF time integration.

Two adaptation experiments are performed from this single base model, each using the predictor-corrector strategy with 100 increments across all layers and a reduced $40\times40=1{,}600$-point collocation grid to bound computation. Two fine-tuning baselines are included: (i) \emph{full-grid} fine-tuning on all $32{,}000$ points and (ii) \emph{sparse-grid} (adapt-grid) fine-tuning on the same $1{,}600$ points used by the projection. Both baselines use adaptive-weight-balanced Adam~\cite{wang2021understanding} followed by L-BFGS, and are run both at the equal wall-clock budget $t_{\text{sub}}$ and to full convergence.

\subsubsection{Initial Condition Adaptation}
\label{sec:burgers_ic}

The first experiment doubles the IC amplitude to $u(x,0)=-2\sin(\pi x)$, retaining the original homogeneous boundaries. The doubled amplitude strengthens the nonlinear convection, steepens the shock, and shifts the shock location. These are fundamentally nonlinear effects that the incremental predictor-corrector strategy captures through continuation rather than a single tangent-space step.

\begin{figure}[ht!]
    \centering
    \includegraphics[width=0.95\textwidth]{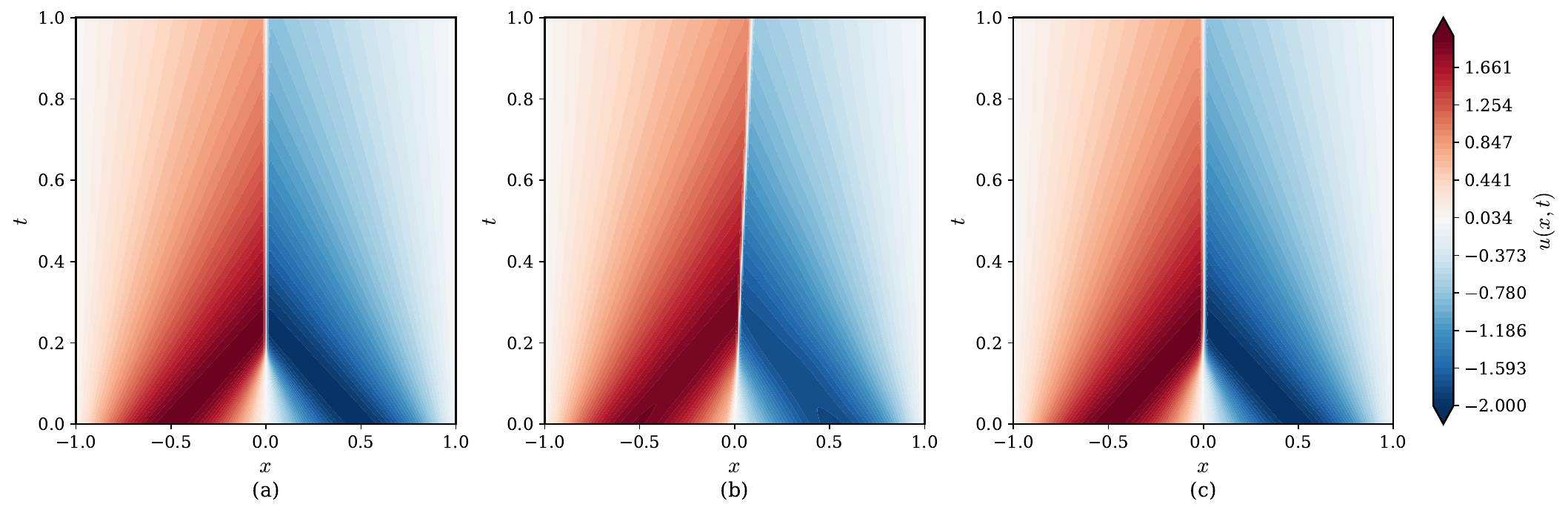}
    \caption{Spatio-temporal contours for the IC adaptation ($u(x,0)=-2\sin(\pi x)$). (a) Reference finite-difference solution. (b) Full-grid fine-tuning: despite achieving very low PDE residual, the solution fails to recover the correct amplitude. (c) Subspace projection: accurate reconstruction of the steepened shock front and doubled amplitude.}
    \label{fig:burgers_ic_adaptation}
\end{figure}

\begin{figure}[ht!]
    \centering
    \includegraphics[width=0.95\textwidth]{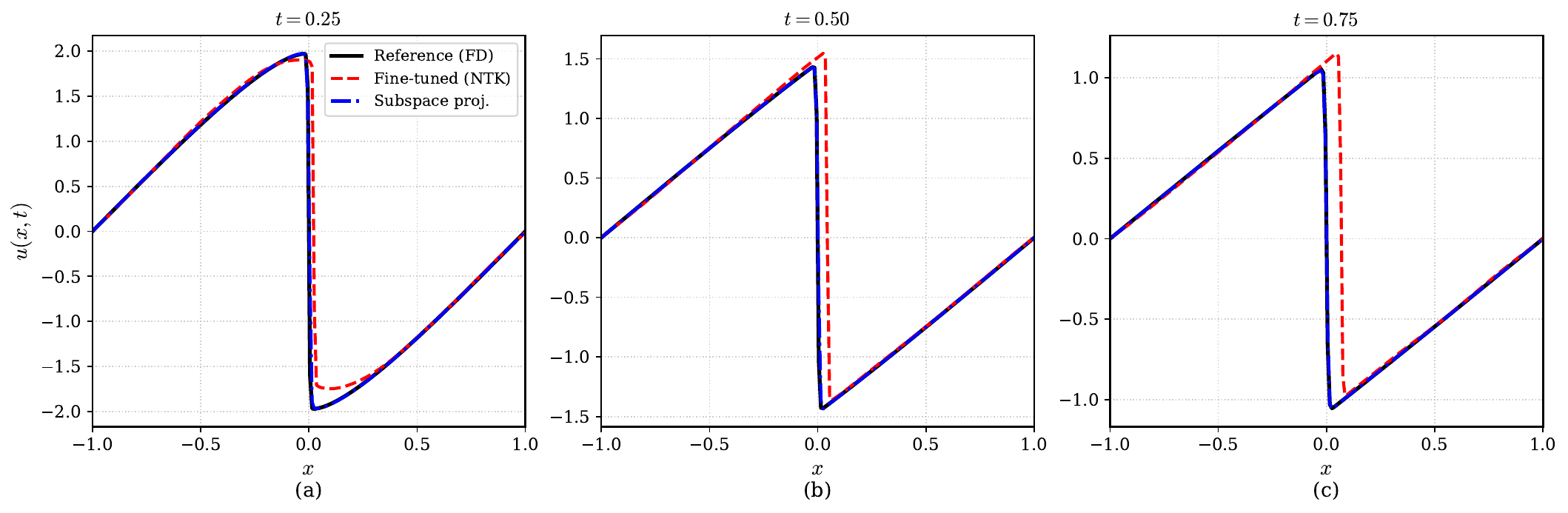}
    \caption{Cross-sectional snapshots of the IC-adapted solutions at $t=0.25$, $0.50$, and $0.75$. Subspace projection (blue) closely tracks the reference shock location and amplitude; full-grid fine-tuning (red) significantly underestimates the solution magnitude.}
    \label{fig:burgers_ic_snapshots}
\end{figure}

Figures~\ref{fig:burgers_ic_adaptation},~\ref{fig:burgers_ic_snapshots} and Table~\ref{tab:burgers_results} compare the adapted solutions against the reference. The IC adaptation is the strongest case for subspace projection: it wins outright on solution quality, not merely on speed. Subspace projection completes in $t_{\text{sub}} = 71.43\,\mathrm{s}$, achieving RMSE $3.91\times10^{-2}$ and enforcing the new IC to a BC MSE of $2.38\times10^{-6}$. Full-grid fine-tuning runs for $1391\,\mathrm{s}$ — nearly $20\times t_{\text{sub}}$ — reaches a full-grid PDE MSE of $1.53\times10^{-5}$ (indistinguishable from the base-model level), yet converges to RMSE $1.02\times10^{-1}$, worse than subspace projection. The network achieves low physics residual but fails to recover the doubled amplitude, as visible in Figure~\ref{fig:burgers_ic_adaptation}(b) and the cross-sectional snapshots of Figure~\ref{fig:burgers_ic_snapshots}. Whether this reflects a shallow local minimum, slow convergence that would eventually improve, or a structural difficulty in separating amplitude from residual under gradient-based optimization, the practical conclusion is the same: subspace projection delivers better solution quality at a fraction of the training cost.

On the $40\times40$ adaptation grid, subspace projection maintains a PDE MSE of $1.54\times10^{-5}$, but $2.29\times10^{1}$ on the full grid. This is expected: the projection respects the physics only on the sparse grid, so the empirically observed null space is larger than the true operator null space, and the PDE residual degrades between the sparse collocation locations. At the equal budget, full-grid fine-tuning has not yet converged (BC MSE $3.64\times10^{-3}$, RMSE $3.58\times10^{-1}$). It reaches the subspace BC target ($2.38\times10^{-6}$) at $t = 175.0\,\mathrm{s}$ ($2.4\times t_{\text{sub}}$) but, as noted, converges to an inferior RMSE. Sparse-grid fine-tuning reaches the same BC target at $t = 118.8\,\mathrm{s}$ ($1.7\times t_{\text{sub}}$) with a full-grid PDE MSE of $1.02\times10^{1}$, confirming that the PDE is not enforced between the sparse points, and converges to RMSE $1.59\times10^{-1}$.

\subsubsection{Boundary Condition Adaptation}
\label{sec:burgers_bc}

The second experiment retains the original IC but imposes time-dependent Dirichlet boundaries $u(-1,t)=0.5t$ and $u(1,t)=-0.5t$. The antisymmetric ramp gradually breaks the spatial symmetry of the base solution, introducing a net positive bias at the left boundary and a growing negative bias at the right.

\begin{figure}[ht!]
    \centering
    \includegraphics[width=0.95\textwidth]{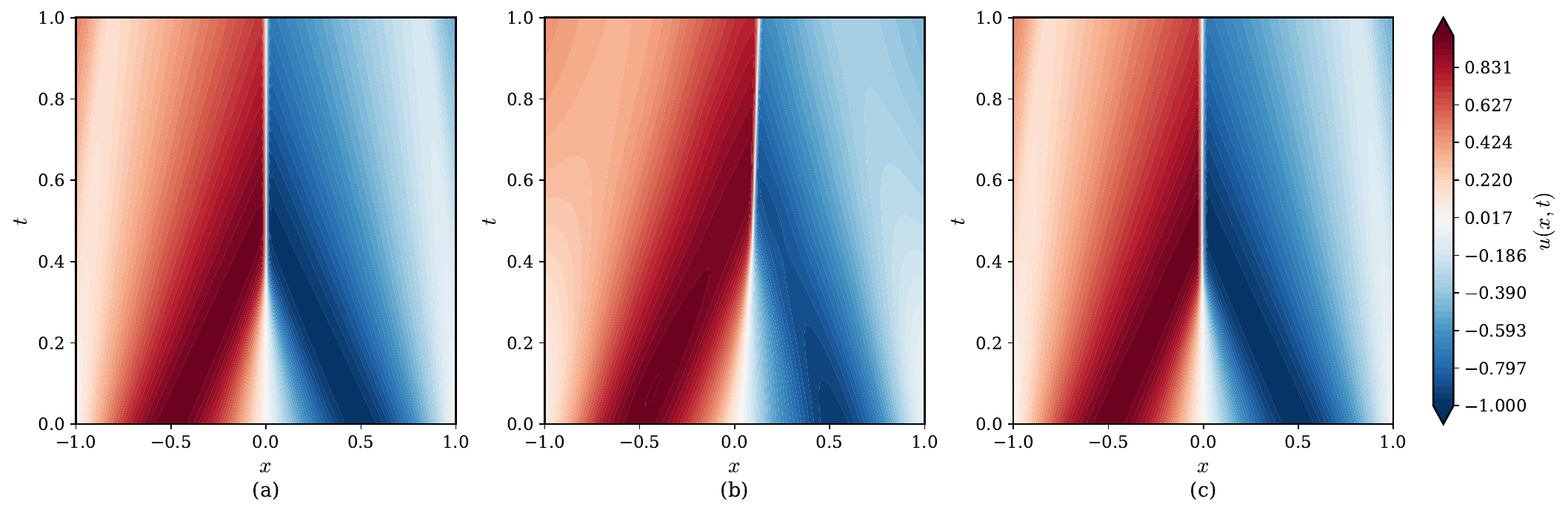}
    \caption{Spatio-temporal contours for the BC adaptation ($u(-1,t)=0.5t$, $u(1,t)=-0.5t$). (a) Reference FD solution with broken spatial symmetry. (b) Full-grid fine-tuning: fails to propagate the new boundary values into the interior within the allotted time. (c) Subspace projection: accurate capture of the asymmetric shock dynamics.}
    \label{fig:burgers_bc_adaptation}
\end{figure}

\begin{figure}[ht!]
    \centering
    \includegraphics[width=0.95\textwidth]{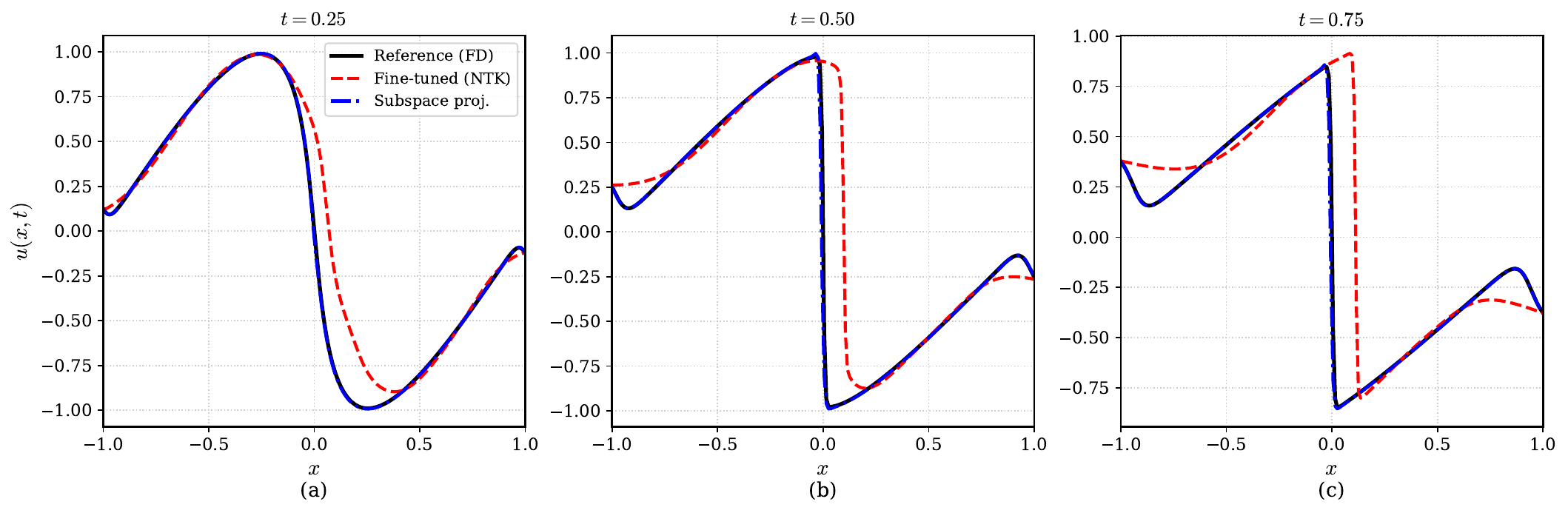}
    \caption{Cross-sectional snapshots of the BC-adapted solutions at $t=0.25$, $0.50$, and $0.75$. Subspace projection (blue) correctly enforces the time-dependent boundary values and resolves the resulting interior asymmetry; fine-tuning (red) fails to match either the boundary values or the interior structure.}
    \label{fig:burgers_bc_snapshots}
\end{figure}

Figures~\ref{fig:burgers_bc_adaptation} and \ref{fig:burgers_bc_snapshots} show the results. Subspace projection completes in $t_{\text{sub}} = 72.30\,\mathrm{s}$, achieving RMSE $4.42\times10^{-2}$ and enforcing the new BCs to a BC MSE of $3.00\times10^{-7}$, while maintaining an adaptation-grid PDE MSE of $5.10\times10^{-5}$ ($7.43\times10^{-1}$ on the full grid). At the equal budget, full-grid fine-tuning has BC MSE $1.19\times10^{-3}$ and RMSE $3.29\times10^{-1}$; sparse-grid fine-tuning fares worse, with BC MSE $3.76\times10^{-2}$ and RMSE $3.78\times10^{-1}$. Full-grid fine-tuning reaches the subspace BC target ($3.00\times10^{-7}$) at $t = 224.3\,\mathrm{s}$ ($3.1\times t_{\text{sub}}$), with a full-grid PDE MSE of $2.58\times10^{-3}$ at the crossing, and converges to RMSE $1.46\times10^{-2}$. Sparse-grid fine-tuning also eventually reaches the BC target at $t = 308.8\,\mathrm{s}$ ($4.3\times t_{\text{sub}}$), but with a full-grid PDE MSE of $8.25\times10^{-1}$, and its final RMSE of $2.51\times10^{-1}$ reflects the cost of enforcing BCs on a sparse grid. Across both scenarios, subspace projection achieves competitive or superior RMSE at a fraction of the time required for fine-tuning to reach comparable BC accuracy, while preserving the physics on the adaptation grid.

\begin{table}[htbp!]
    \centering
    \caption{Burgers equation adaptation results. The base model PDE MSE is $1.18\times10^{-6}$. Adapt-grid PDE MSE is evaluated on the $40\times40$ adaptation grid (subspace projection only); full-grid PDE MSE on the $200\times160$ grid; RMSE against the finite-difference reference on a $200\times200$ grid. Fine-tuning results are reported at equal budget ($t_{\text{sub}}$) and at training end.}
    \label{tab:burgers_results}
    \resizebox{\textwidth}{!}{
        \begin{tabular}{llccccc}
            \toprule
            \textbf{Scenario} & \textbf{Method}                & \textbf{Time (s)} & \textbf{Adapt PDE MSE} & \textbf{Full PDE MSE} & \textbf{BC MSE}      & \textbf{RMSE}       \\
            \midrule
            \multirow{5}{*}{IC adapt}
                              & Subspace proj.                 & $71.43$           & $1.54\times10^{-5}$    & $2.29\times10^{1}$    & $2.38\times10^{-6}$  & $3.91\times10^{-2}$ \\
                              & Full FT (at $t_{\text{sub}}$)  & $71.43$           & ---                    & $2.21\times10^{-2}$   & $3.64\times10^{-3}$  & $3.58\times10^{-1}$ \\
                              & Full FT (at end)               & $1391.44$         & ---                    & $1.53\times10^{-5}$   & $2.37\times10^{-10}$ & $1.02\times10^{-1}$ \\
                              & FT adapt (at $t_{\text{sub}}$) & $71.43$           & ---                    & $1.74\times10^{1}$    & $3.94\times10^{-5}$  & $2.13\times10^{-1}$ \\
                              & FT adapt (at end)              & $1057.52$         & ---                    & $1.99\times10^{1}$    & $1.39\times10^{-9}$  & $1.59\times10^{-1}$ \\
            \midrule
            \multirow{5}{*}{BC adapt}
                              & Subspace proj.                 & $72.30$           & $5.10\times10^{-5}$    & $7.43\times10^{-1}$   & $3.00\times10^{-7}$  & $4.42\times10^{-2}$ \\
                              & Full FT (at $t_{\text{sub}}$)  & $72.30$           & ---                    & $7.15\times10^{-2}$   & $1.19\times10^{-3}$  & $3.29\times10^{-1}$ \\
                              & Full FT (at end)               & $1439.48$         & ---                    & $6.42\times10^{-6}$   & $6.57\times10^{-9}$  & $1.46\times10^{-2}$ \\
                              & FT adapt (at $t_{\text{sub}}$) & $72.30$           & ---                    & $1.84\times10^{-2}$   & $3.76\times10^{-2}$  & $3.78\times10^{-1}$ \\
                              & FT adapt (at end)              & $1008.18$         & ---                    & $8.32\times10^{-1}$   & $1.65\times10^{-7}$  & $2.51\times10^{-1}$ \\
            \bottomrule
        \end{tabular}
    }
\end{table}

\section{Conclusion}
\label{sec:conclusion}

This work develops a framework for analyzing and adapting physics-informed neural networks via the Fisher Information Matrix. We introduced an effective-dimension metric $d_{\text{eff}}$ that quantifies the parameter directions left unconstrained by the governing PDE, used it as a structural diagnostic to detect under-resolution that training loss can hide, and exploited the null space it identifies to adapt boundary conditions without disturbing the learned physics in exact arithmetic, with finite-precision drift controlled by a predictor-corrector retraction.

The empirical findings of this work are threefold. First, for finite-kernel operators, $d_{\text{eff}}^{\text{pde}}$ converges to the analytical kernel dimension of the differential operator, independent of network width, depth, or activation, across linear ODEs of order 1--3, the Bessel operator, and nonlinear second-order operators; for the 2D Poisson operator, whose kernel is infinite-dimensional, $d_{\text{eff}}$ measures representational bandwidth rather than recovering an integer invariant (Section~\ref{sec:invariance}). Second, in the under-resolved regime the metric saturates the collocation-point rank ceiling, providing a structural diagnostic for whether the network has absorbed the operator's geometric content (Section~\ref{sec:invariance}). Third, the null space identified by the metric enables boundary adaptation with minimal tuning overhead: subspace projection reaches near-equivalent boundary accuracy 3--34$\times$ faster than gradient-based fine-tuning reaches comparable BC accuracy, requiring no per-problem loss weighting or optimizer scheduling (Sections~\ref{sec:subspace_strategy}--\ref{sec:multi_dimensional}). Gradient-based fine-tuning can achieve comparable or better final accuracy at full convergence; the advantage of subspace projection is the elimination of that convergence wait.

Subspace projection removes the physics-boundary loss tradeoff during adaptation by confining parameter updates to the physics null space; gradient-based fine-tuning resolves the same tradeoff through optimization, which can match or exceed final solution quality at the cost of additional convergence time and per-problem tuning. Finite-precision arithmetic and the curvature of the parameter-to-output map introduce small but unavoidable residual drift into the projection. We control this drift via a predictor-corrector retraction that projects accumulated errors back to the physics null space at each step. The remaining limitations are the quality of the local linearization used in each Gauss-Newton step and the computational budget allocated to adaptation.

The function-space reformulation (Section~\ref{sec:subspace_strategy}) addresses the dominant $\mathcal{O}(d^3)$ parameter-space cost, bringing computation into the $\mathcal{O}((N_p + N_b)^3)$ regime. For very large numbers of collocation or boundary points, further reduction may be possible via low-rank truncation of $VV^\mathsf{T}$: retaining only the top-$k$ eigenvectors reduces the cost to $\mathcal{O}(kd^2)$. We do not pursue this optimization in the present work but note it as a natural direction for scaling to industrial-scale problems.

The framework supports a workflow well-suited to scientific computing: pre-train a base model entirely on the PDE residuals to capture the physical manifold across the domain, then post-hoc anchor any subsequently observed boundary conditions or sparse data points via subspace projection. The base model represents the physics; the projections enforce the constraints. This is complementary to operator learning approaches~\cite{lu2021learning,li2020fourier,wang2021learning}, which solve a related but distinct problem (in-distribution boundary adaptation given an offline training distribution). Our method handles adaptation including boundary conditions outside any pre-specified distribution, making the two approaches natural choices for different regimes.

Future work includes an extension to neural architecture search, removing all parameters that lie in the null space of the combined PDE and BCs and extension of the low-rank truncation strategy for industrial-scale problems. Beyond boundary adaptation, the null-space structure also opens a path to physics-constrained uncertainty quantification from a single trained model.

\section*{CRediT Authorship Contribution Statement}

\textbf{Cornelius Otchere:} Conceptualization, Methodology, Software, Formal analysis, Investigation, Writing (original draft).
\textbf{Michael Shields:} Conceptualization, Supervision, Funding acquisition, Writing (review \& editing).

\section*{Acknowledgments}

Cornelius Otchere thanks Professor James C. Spall, whose reading assignment in System Identification and Likelihood Methods inspired this paper.

\section*{Data Availability}

The code used to generate the results in this paper will be made publicly available upon acceptance.

\section*{Declaration of Competing Interests}

The authors declare that they have no known competing financial interests or personal relationships that could have appeared to influence the work reported in this paper.

\section*{Declaration of Generative AI and AI-Assisted Technologies in the Writing Process}

During the preparation of this work the authors used Claude for editing and formatting assistance. After using this tool, the authors reviewed and edited the content as needed and take full responsibility for the content of the published article.

\bibliography{bibliography}
\end{document}